\documentclass{article}

\usepackage{microtype}
\usepackage{graphicx}
\usepackage{subfigure}
\usepackage{booktabs} 

\usepackage{booktabs}

\usepackage{needspace}

\usepackage{hyperref}

\usepackage[table,xcdraw]{xcolor}

\usepackage[accepted]{icml2025}

\usepackage{amsmath}
\usepackage{amssymb}
\usepackage{mathtools}
\usepackage{amsthm}

\usepackage{listings}
\usepackage{inconsolata}

\newlength{\jsondescindent}
\lstdefinestyle{icmljson}{
  basicstyle=\ttfamily\footnotesize,
  columns=fullflexible,
  keepspaces=true,
  showstringspaces=false,
  frame=none,
  breaklines=true,
  breakatwhitespace=true,
  breakautoindent=false,      
  breakindent=5em,
  xleftmargin=0pt,
  xrightmargin=0pt,
  linewidth=\linewidth        
}

\usepackage[most]{tcolorbox}
\tcbset{
  colback=gray!5,   
  colframe=black!15, 
  boxrule=0.3pt,   
  arc=2pt,         
  left=4pt, right=4pt, top=4pt, bottom=4pt 
}

\usepackage[capitalize,noabbrev]{cleveref}

\theoremstyle{plain}

\theoremstyle{definition}

\theoremstyle{remark}

\usepackage[textsize=tiny]{todonotes}

\icmltitlerunning{Natural Language Tools: A Natural Language Approach to Tool Calling In Large Language Agents}

\begin{document}

\twocolumn[
\icmltitle{Natural Language Tools: A Natural Language Approach to Tool Calling In Large Language Agents}



\newcommand{\authorspacing}{3em} 

\vspace{0.3em}
\centerline{\textbf{Reid T.\ Johnson}\hspace{\authorspacing}\textbf{Michelle D.\ Pain}\hspace{\authorspacing}\textbf{Jordan D.\ West}}
\vspace{0.8em}
\centerline{PokketCoach}
\vspace{0.5em}

\renewcommand{\thefootnote}{}
\setcounter{footnote}{0}


\icmlkeywords{Machine Learning, ICML}

\vskip 0.3in
]

\begin{abstract}
We present Natural Language Tools (NLT), a framework that replaces programmatic JSON tool calling in large language models (LLMs) with natural language outputs. By decoupling tool selection from response generation, NLT eliminates task interference and format constraints that degrade tool call performance. When evaluated across 10 models and 6,400 trials spanning customer service and mental health domains, NLT improves tool calling accuracy by 18.4 percentage points while reducing output variance by 70\%. Open-weight models see the largest gains, surpassing flagship closed-weight alternatives, with implications for model training in both reinforcement learning and supervised fine-tuning stages. These improvements persist under prompt perturbations and extend tool-calling capabilities to models lacking native support.
\end{abstract}

\section{Introduction}
\label{sec:intro}

Large language models (LLMs) have dramatically expanded their capabilities through tool calling, the ability to invoke external functions such as web search, information retrieval, and code execution \citep{Qu2024ToolUse, Li2024ToolUse}. Beginning with early systems like WebGPT \citep{Nakano2021WebGPT}, these capabilities have become foundational to modern agentic architectures, enabling LLM-based agents to interface with other systems to orchestrate complex actions. A variety of benchmarks have been developed to measure tool calling efficacy \citep{Huang2024ToolUtilizationBenchmark, Yao2024TaoBench}, while model providers consistently highlight even modest improvements in accuracy \citep{OpenAI2025GptOss, Anthropic2025Sonnet45}. However, even state-of-the-art models frequently fail to make accurate tool calls \citep{Maekawa2025Tools}.

Current implementations typically require models to emit structured outputs conforming to rigid function schemas (i.e., JSON or XML). We refer to this as ``structured tool calling.'' Figure~\ref{fig:json} demonstrates such an output.

\begin{figure}[h]
\centering
\begin{minipage}{0.95\linewidth}
\begin{lstlisting}[style=icmljson,
  label={fig:json}]
{
  "tool_calls": [
    {
      "name": "check_talk_to_a_human",
      "description": "Used when the user requests to speak with a human representative, agent, or support person. This tool checks whether escalation to a human is possible and initiates the handoff process if appropriate."
    }
  ]
}
\end{lstlisting}
\end{minipage}
\caption{An example of a JSON-formatted tool call for a ``talk to a human'' request within a customer service scenario.}
\label{fig:json}
\end{figure}

This structured approach has become widely adopted, beginning with releases such as Toolformer \citep{Schick2023Toolformer} and extending to most major model providers \citep{OpenAI2023ToolAnnounce, Google2023ToolAnnounce, Anthropic2024ToolAnnounce}. Yet as tool calling becomes an increasingly critical component of LLM-powered agents, growing evidence suggests the underlying structure may come with significant drawbacks \citep{Gupta2024TaskInterference, Tam2024LetMeSpeakFreely, Wang2025SLOT}. Structured formats require models to simultaneously handle multiple competing demands, such as: understanding the query, selecting appropriate tools, adhering to format constraints, and generating a response. \citet{Gupta2024TaskInterference} demonstrated that such task interference can degrade performance significantly, with some models experiencing more than 20\% reductions in accuracy. \citet{Tam2024LetMeSpeakFreely} provided another striking example: requiring JSON output reduced response accuracy by 27.3 percentage points on the GSM8K benchmark compared to natural language. Tam et al. further reported that the more constrained a model's output, the more pronounced this loss of accuracy.

Beyond format constraints alone, structured tool definitions increase context length \citep{Paramanayakam2024LessisMore}. As demonstrated by \citet{Levy2024SameTaskMoreTokens} and \citet{Modarressi2025NoLiMa}, an increase in input tokens rapidly degraded model performance well before its maximum context window. Modarressi et al. found that some models experienced 16 percentage point accuracy reductions when context length increased by as little as 1,000 tokens, and reductions of up to 50 percentage points as context length surpassed 8,000 tokens. This prompts the question: is the structured tool calling approach a bottleneck for performance?

Recent efforts have sought to minimize these performance degradations. \citet{Wang2025SLOT} demonstrated that architectural changes can bolster accuracy by more than 20 percentage points. Their work achieved this by passing an unstructured output through a second model, transforming the output into an appropriately structured format. Further research by \citet{Chen2024EnhancingFunctionCalling} addressed tool calling directly, crafting an intermediate step where a second model selects the appropriate tools from a list and passes them in structured format to the responding model. Other approaches have prompted the model with specific thinking instructions to improve tool call accuracy \citep{Dang2025GuidedStructured} or worked to compress the definitions of tool calls to minimize token use \citep{Yuan2024Compression}. While these methods improved reliability, they remained within the structured tool calling paradigm rather than questioning whether those structures were necessary for tool calling at all.

We propose Natural Language Tools (NLT), which decouples tool selection into a dedicated model step while departing from programmatic output structures entirely. Instead of emitting JSON, models list each available tool along with a simple YES or NO decision. A parser then executes the selected tools, passing them to a final response module. This architecture isolates tool calling to its own discrete component, reducing both task interference and token overhead.

We evaluate this approach across 10 frontier models, spanning 6,400 trials in customer service and mental health domains. We find that NLT improves tool calling accuracy from 69.1\% to 87.5\%, an 18.4 percentage point gain compared to structured baselines. Open-weight models benefit the most (+26.1 percentage points), though flagship closed-weight models still see significant lift (+10.6 percentage points). These findings suggest that natural-language implementations of tool calling can significantly boost agentic performance compared to rigidly structured alternatives. All prompts and inputs can be found in Appendix A.

\section{NLT Implementation}
\label{sec:implementation}

Natural Language Tools (NLT) represents a modular framework for tool calling, separating tool selection from response generation while simultaneously departing from programmatic structures. By eliminating programmatic format constraints and isolating tool selection, NLT addresses both the task interference and token overhead challenges identified in Section~\ref{sec:intro}. This section details the core architecture, prompt design principles, and interface design.

\subsection{NLT System Architecture}
\begin{figure*}[t]
\centering
\includegraphics[width=0.95\textwidth]{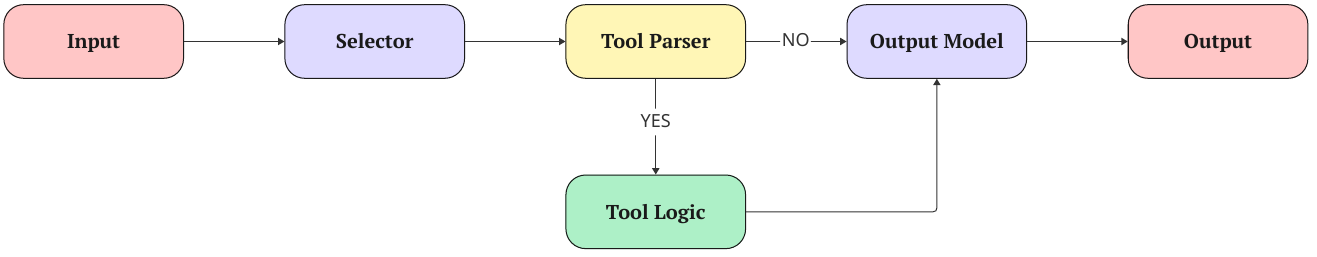}
\caption{The above represents a simplified NLT architecture with a single input LLM and a single output LLM. As NLT is fully modular, these layers can be abstracted to include any level of additional complexity. For example, one tool may be chained into another, or the final output model may simply be the next step in an agentic process.}
\label{fig:arch}
\end{figure*}

NLT is a modular three-step architecture that separates tool selection from response generation, illustrated in Figure~\ref{fig:arch}.

\paragraph{Stage 1: Tool Selection.}
A \emph{selector} model receives an input along with a crafted NLT prompt. Following prior work \citep{Wei2023COT, Yao2023ReAct, Masterman2024ToolReasoning}, the model is instructed to think through whether any tools are relevant, then output each available tool within its response along with a \texttt{YES} or \texttt{NO} determination. This format allows multiple tools to be called simultaneously. This output is then passed to a parser.
\paragraph{Stage 2: Tool Execution.}
A parser determines the status of each tool from the selector output. The simple \texttt{YES} or \texttt{NO} format and consistent language enables straightforward parsing using regex or string matching. If a tool is labeled \texttt{YES}, the system executes any logic or functions relevant to that tool.
\paragraph{Stage 3: Response.}
After the relevant tools have been executed, NLT continues to a subsequent output module. This module may involve another LLM responding as if in a conversation, or may lead to some other systemic process. The NLT architecture simply acts as an intermediary layer between the initial input and this output module.

\subsection{Selector Prompt Structure}
The selector system prompt is a core component of NLT, as it constrains the Selector Model's output and defines its ``toolbox.'' We design a five-component template to provide clear context and consistent output formatting:
\begin{enumerate}
  \item \textbf{\{NLT-role\}.} A short description of the context of the conversation, and the role the selector plays in that context. This grounds the model in its role within the context of the agent as a whole.
  \item \textbf{\{Search-purpose\}.} A short overall description of the explicit goals for the selector. This explicitly establishes the tool-calling motive for the selector.
  \item \textbf{\{Tool-list\}.} A list of each available tool along with a description of when to use the tool, both in natural language. This gives the selector key information on when and how to use each tool.
  \item \textbf{\{Output-description\}.} A description of the desired output format in natural language. This establishes the desired parsable format for the model to follow, without resorting to programmatic restrictions.
  \item \textbf{\{Output-example\}.} An explicit example of the desired output, detailing every tool call with an explicit \texttt{YES}/\texttt{NO} option. This further incentivizes the correct output format.
\end{enumerate}

\subsection{Selector Prompt Example}
The following example illustrates the complete NLT prompt structure for a hypothetical customer service scenario. This exemplifies the core principle of NLT: replacing programmatic constraints with natural language to guide tool selection. 
\begin{tcolorbox}
\small You are an assistant to Messenger, a chatbot designed to help with customer service requests.\\[0.4em]

Your mission is to identify if any of the following topics have been brought up or if an action is relevant.\\[0.4em]

Your list of possible options include the following:\\[0.4em]

Website Information (questions about your website: where to purchase tickets or merch, order tracking, policies, or how to use the site)\\[0.4em]
Talk to a Human (explicit requests to reach a live agent by chat/phone/email)\\[0.4em]
Past Purchases (questions about previous orders, receipts, warranties, returns, or order lookups)\\[0.4em]

Your output should begin by thinking whether or not any of these are brought up or if they are relevant. Then, include the name of every topic, followed by either yes or no. You should end with ``Assessment finished.''\\[0.4em]

It should always be in the following format:\\[0.4em]

\texttt Thinking: (insert\_thinking)\\[0.4em]
\texttt Website Information -- YES/NO\\[0.4em]
\texttt Talk to a Human -- YES/NO\\[0.4em]
\texttt Past Purchases -- YES/NO\\[0.4em]
\texttt Assessment finished.
\end{tcolorbox}

\section{Methods}
To evaluate NLT's effectiveness compared to structured tool calling, we designed a comprehensive comparative study across multiple model families and domains. We focused on single-turn, parameterless tool calling to enable clear comparison between approaches while using prompt perturbations to control for prompt engineering effects.

\subsection{Protocol}
We evaluated performance using a 2 $\times$ 2 $\times$2 factorial design, crossing approach (Structured Tool Calling vs NLT), scenario (Customer Service vs Mental Health), and perturbation (Non-perturbed vs Perturbed). Each scenario contained 16 user inputs designed to cover the full range of tool calling demands. Some inputs required no tools, others required a single tool, and still others required parallel selection of multiple tools simultaneously. These same 16 inputs were used across both non-perturbed and perturbed prompts to isolate the impact of prompt variations.

Each trial consisted of a single API call evaluating one user input under specific experimental conditions, with no conversation history or context carryover between trials. We employed strict exact-match evaluation: a trial succeeded if and only if the model's tool selection exactly matched the predetermined ground truth. No partial credit was granted, with extra or missing tools resulting in binary pass/fail scoring. This binary approach avoided ambiguity in comparing approaches while setting a high bar for success.

To account for variation in model outputs, we ran five independent replications of each input under default settings. Each replication invoked a fresh API call to the model provider and ran in parallel, ensuring true independence rather than re-parsing of cached responses. This methodology yielded 640 trials per model (2 approaches $\times$ 2 scenarios $\times$ 16 inputs $\times$ 2 prompt versions $\times$ 5 replications). Across our 10-model core set, this produced 6,400 trials. For each trial, we logged the raw model output, parser results, and total token usage. API errors, rate limits, and timeouts were excluded from analysis and retried until clean responses were obtained.

\subsection{Test Scenarios and Ground Truth}
Models vary widely in their capabilities depending on the domain \citep{Afzal2024AdaptEval,Kim2025BenchHub}. To address this and provide a thorough comparison between structured tool calling and NLT, we tested two distinct scenarios. These were chosen due to their broad popularity as use cases for LLMs \citep{Raza2025IndustrialApplications}, while still providing a complex enough landscape to stress single-turn parallel tool calling capabilities.

\subsubsection{Scenario 1: ``Alex'' (Customer Service)}
Customer service agents remain one of the primary use cases of consumer-facing LLMs \citep{IBM2024CustomerService}. In this scenario, Alex acts as a customer support agent for a fictional music venue named ``Yes! Music''. Seven tools were chosen for this, covering memory functions, information retrieval, customer account access, and transfers to human staff.

\subsubsection{Scenario 2: ``Sage'' (Mental Health)}
Interest in language models for mental health support has existed since the advent of LLMs \citep{Hua2025LLMmentalHealth}, and stretches back even to the origins of AI systems \citep{Weizenbaum1966ELIZA}. In this scenario, Sage acts as a mental health support agent, representing a more specialized domain with safety considerations. Eight tools were selected covering assessment functions, information access, boundary management, and critical safety interventions.  \emph{Complete tool lists and descriptions appear in Appendix~A.}

\subsubsection{Ground Truth Determination}
Each scenario was assessed via 16 user inputs, covering a variety of possible topics and requests. These inputs were crafted to represent realistic messages, including spelling mistakes and emotionally charged messages, while also addressing a comprehensive cross-section of the available tools. They are not based on excerpts from real conversations, but are fully synthetic.

Of the 16 inputs in the Alex scenario, 2 required zero tools, 7 required single tools, and 7 required parallel tool selection.
Of the 16 inputs in the Sage scenario, 4 required zero tools, 5 required single tools, and 7 required parallel tool selection.

Expected tool calls for each input were determined before trials began. Given the safety-critical nature of the mental health scenario, tool determinations involving potential self-harm or crisis interventions were independently validated by a licensed psychologist with extensive experience in mental health assessment.

\subsection{Models}
We evaluated 13 models spanning both open-weight and closed-weight families. Models were selected based on popularity at the time of this study via the OpenRouter leaderboard  \citep{OpenRouter2025}, along with the inclusion of topical open-weight models such as OpenAI's OSS family and Qwen.

Our primary analysis assesses 10 distinct models with access to parallel tool-calling. 

\textbf{Like-for-like core set.}
OpenAI gpt-5; OpenAI gpt-5-nano; Google gemini-2.5-pro; Google gemini-2.0-flash; Anthropic claude-sonnet-4-20250514; Meta Llama-4-Maverick-17B-128E-Instruct-FP8; Meta Llama-4-Scout-17B-16E-Instruct; Moonshot Kimi-K2-Instruct; Alibaba Qwen3-235B-A22B-Thinking-2507; DeepSeek DeepSeek-V3-0324.

Three additional models were not tested on structured approaches due to limited tool calling capabilities at evaluation time. Instead, their NLT results are examined to assess how this approach can extend model capabilities.

\textbf{Auxiliary set (non-comparable at evaluation time).}
DeepSeek R1-0528, OpenAI GPT-OSS-120B, GPT-OSS-20B.

This selection ensures a diverse set of flagship models is covered from multiple model families and providers.

\subsection{NLT vs. Structured Tool Calls}
We evaluated two approaches to tool calling:

\textbf{1) built-in structured tool calling approaches within each model.} The model received a system prompt containing the tool descriptions and names, tool descriptions (as structured functions sent via the model host API), and the user input; and

\textbf{2) the NLT approach.} The model received an NLT-structured system prompt containing identical tool descriptions to the structured tool calling counterpart, and the user input.

Structured and NLT approaches necessarily differ in system prompts: structured approaches define their tool calls with API-level function formatting, while NLT uses natural language descriptions within the prompt. To ensure fair comparison, as much similarity was preserved as possible, with identical tool descriptions, role descriptions, user inputs, and grading criteria. \emph{See Appendix~A for exact system prompts and tool descriptions.}

\subsection{Perturbations}
Minor changes in system prompts can lead to major shifts in model performance \citep{Sclar2024PromptSensitivity, Errica2025PromptSensitivity}. Experienced practitioners can use this to their advantage, significantly boosting model performance through intentionally designed prompts \citep{Brown2020fewshot, Kojima2023ZeroShotReasoners}. To account for the impact of prompt engineering skill on tool calling performance and test NLT robustness, a second version of each scenario was maintained. In this second variation, all tool descriptions and system prompts were perturbed. User inputs were not perturbed during trials.  

Each tool description and system prompt was given to OpenAI's gpt-5-nano, with instructions to rewrite the content while maintaining overall purpose and structure. This approach allowed us to automate perturbations without introducing prompt engineering bias from a human prompter. Tool descriptions were assessed post-perturbation to ensure they still accurately reflected intended use, with a full list of perturbations available in Appendix~A.

\subsection{Data Collection}
We accessed all models via API: open-weight models via the TogetherAI platform \citep{TogetherAILink}, and closed-weight models through their respective provider APIs. For each trial, we recorded raw model outputs, parser results, and token usage. API errors, rate limits, and timeouts were discarded and retried until clean responses were obtained.

\section{Results}
This section reports comparative performance between the NLT interface and the structured tool calling (i.e. JSON) baseline. Unless otherwise stated, results are aggregated over the primary analysis set of ten like‑for‑like models with parallel tool calling.

\subsection{Overall Accuracy}
\begin{figure}[t]
    \centering
    \includegraphics[width=\columnwidth]{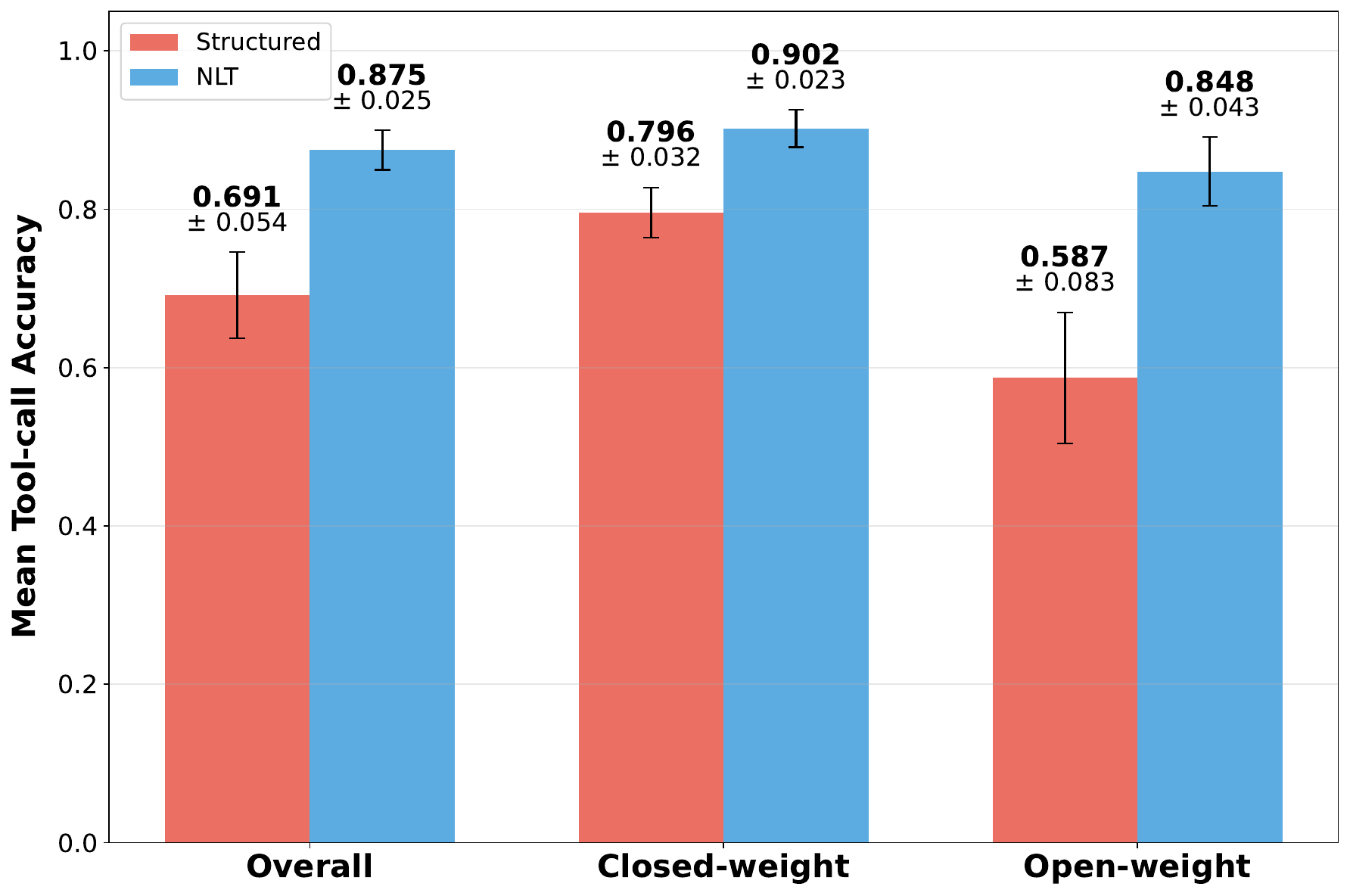}
    \caption{Side by side comparisons between NLT and Structured approaches, across three categories: overall results, closed-weight models, and open-weight models.}
    \label{fig:overall_performance}
\end{figure}

We assess overall tool-call accuracy across all models and scenarios through like-for-like comparisons of overall accuracy in both approaches. We report these results in Figure~\ref{fig:overall_performance}.

Across all models and scenarios, NLT improves tool‑call accuracy by 18.4 percentage points on average relative to the structured baseline, an increase from 69.1\% to 87.5\%. Improvements are observed across all model families and task scenarios. Open‑weight models showcase larger gains (+26.1 percentage points, 58.7\% to 84.8\%) compared to their closed‑weight counterparts (+10.6 percentage points, 79.6\% to 90.2\%).

\subsection{Per-model Performance}

\begin{figure}[t]
    \centering
    \includegraphics[width=\columnwidth]{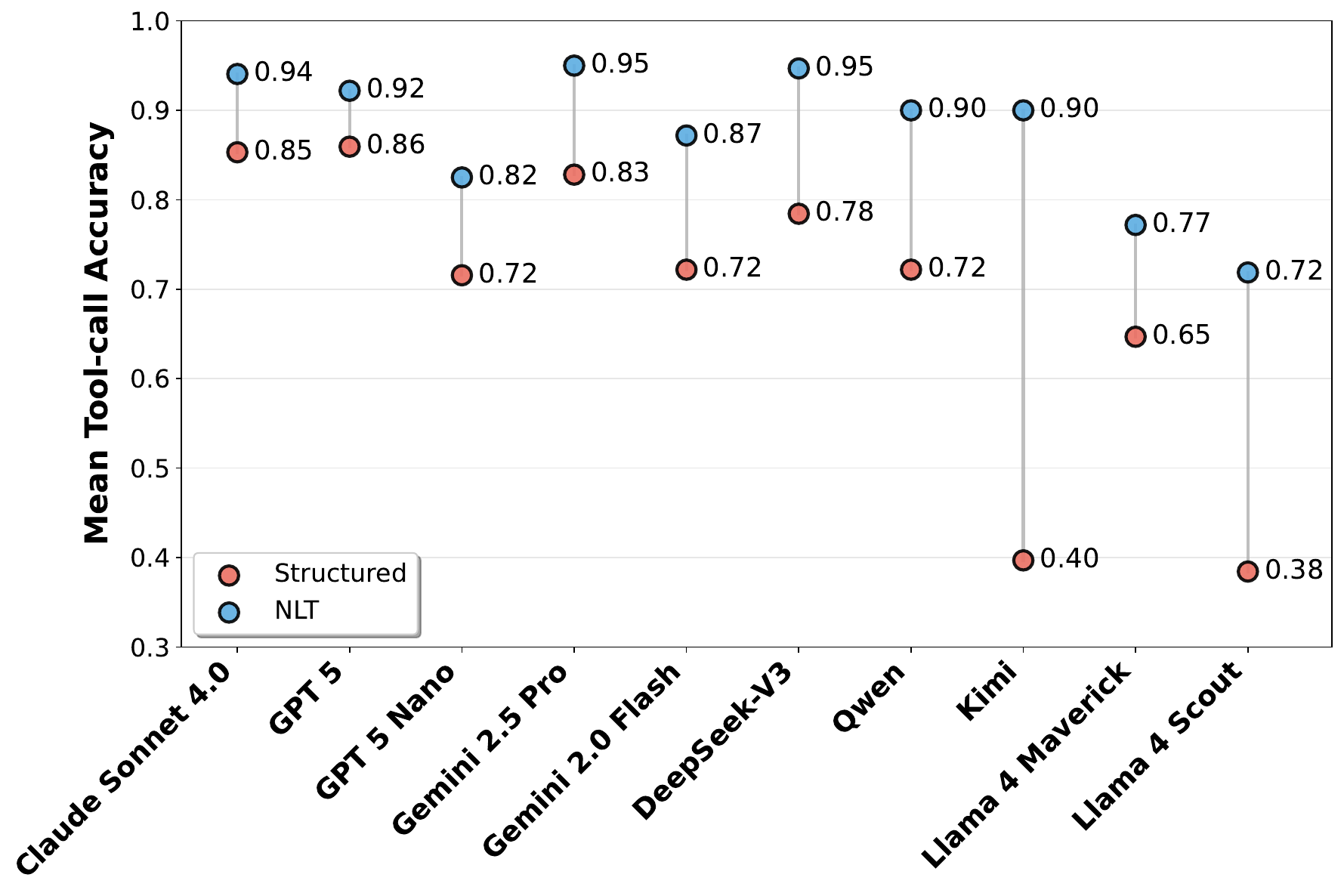}
    \caption{Relative comparisons between Structured and NLT tool-calling approaches across core model set. Mean tool-call accuracy is averaged across domains and perturbations.}
    \label{fig:permodel}
\end{figure}

Different large language models may have widely different capabilities, even within the same model family \citep{Ye2023CapabilityGPT3, Afzal2024AdaptEval, Kamalov2025EntrepreneurshipChatbots}. To help ensure NLT is not an artifact of model-specific weights or training processes, we assess results on a per-model basis. We report these results in Figure~\ref{fig:permodel}.

Per‑model tool call accuracy favors NLT for all assessed models. However, model-specific improvements vary widely. For example, OpenAI's GPT-5 displays a 6.2 percentage point increase in tool calling accuracy using NLT. Yet Gemini 2.5 Pro nearly doubles this, with a 12.2 percentage point lift. Other gains are more pronounced; Moonshot's Kimi K-2 Instruct sees a boost of more than 50 percentage points. 

\begin{figure*}[t]
\centering
\includegraphics[width=\textwidth]{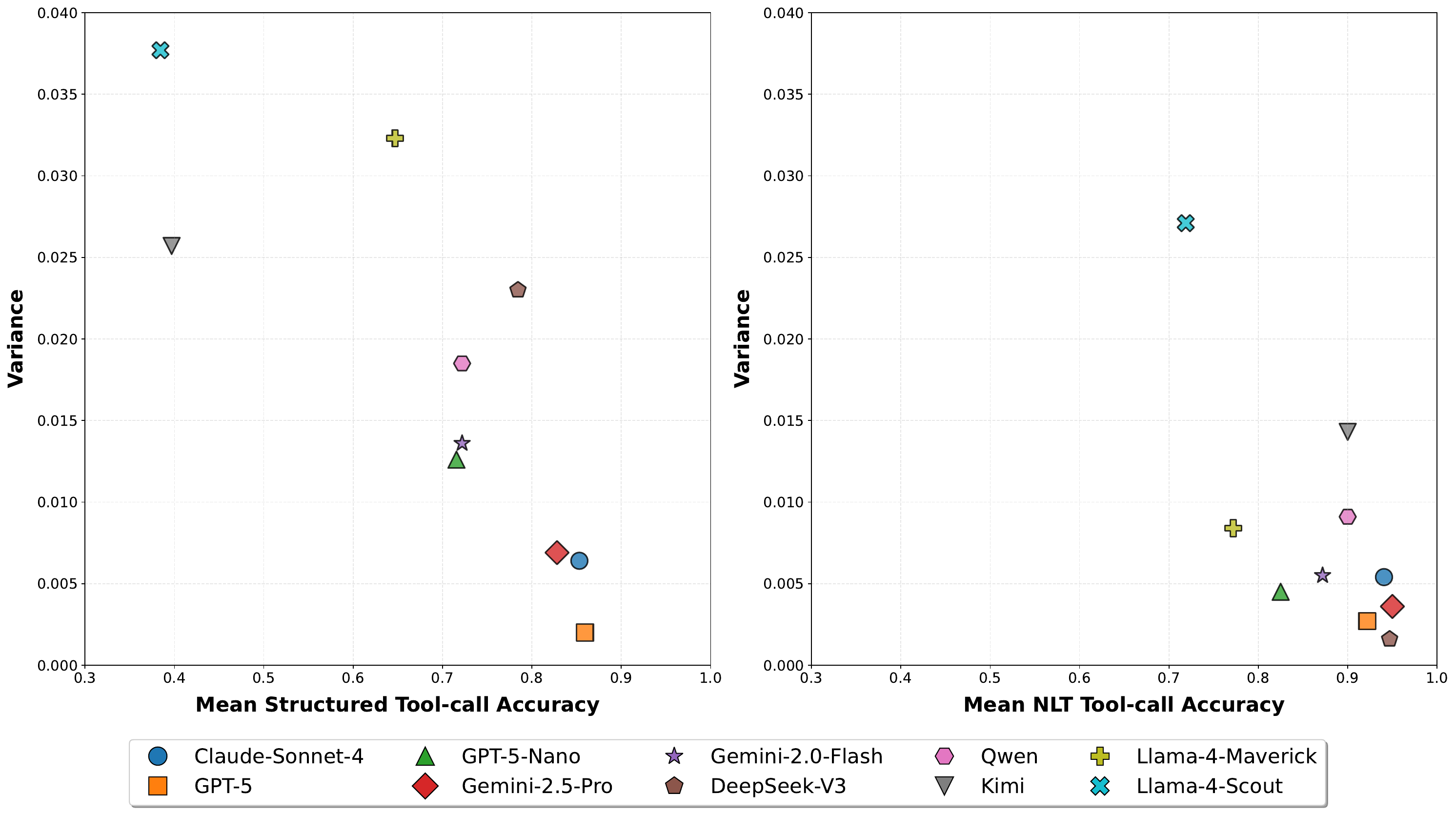}
\caption{Side by side comparisons of model performance shifts between structured tool calls and NLT, with variance represented by the y-axis and mean accuracy by the x-axis. Accuracy and variance are calculated across the aggregate of Alex and Sage scenarios, as well as perturbed and non-perturbed prompts. Full per-model results are reported in Appendix~B.}
\label{fig:variance}
\end{figure*}

Beyond increasing accuracy, reducing variance in outputs is critical in production systems where inconsistent tool calling could cause failures. We examine both accuracy and variance across models in Figure~\ref{fig:variance}.

\pagebreak
Model variance is significantly reduced under NLT compared to structured tool calling. Structured tool calling achieves a mean accuracy of 69.1\% with high variability (variance = 0.0411, SD = 20.28 percentage points). NLT achieves marked improvements in both metrics, with 87.5\% mean accuracy and much lower variability (variance = 0.0121, SD = 10.99 percentage points).

Nearly all models see lower variance under NLT (Appendix B). The sole exception is OpenAI's GPT-5, which sees increased variance from 0.0020 to 0.0027. However, this increase in variance is offset by significantly higher accuracy (85.9\% with structured compared to 92.2\% with NLT). 

This suggests that NLT not only improves average accuracy, but also provides more predictable and consistent tool calls, a critical capability for agentic systems in the real world. DeepSeek V3 dramatically demonstrates this improvement, increasing mean accuracy from 78.4\% to 94.7\% while simultaneously reducing variance from 0.023 to 0.0016, going from among the least stable performers to the most consistent model. 

\subsection{Robustness to Prompt Perturbation}

\begin{figure}[t]
\centering
\includegraphics[width=\columnwidth]{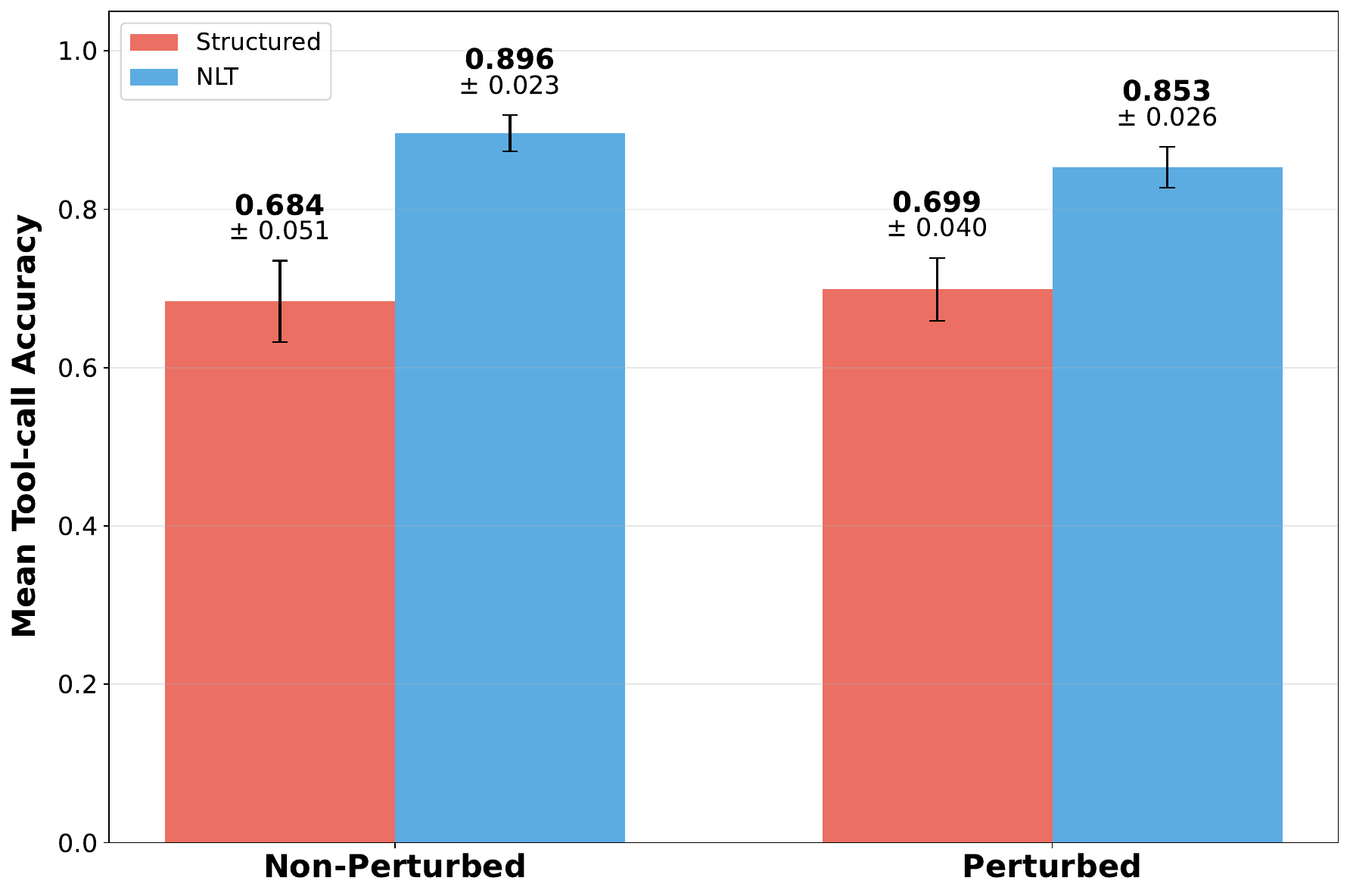}
\caption{A comparison of non-perturbed conditions to perturbed, demonstrating their performance shifts between structured and NLT tool-calling approaches.}
\label{fig:perturbations}
\end{figure}

We additionally assess the NLT approach in an alternative set of system prompts and tool definitions, perturbed to reflect input variance. Performance improvements persist even when system prompts and tool descriptions are perturbed. We report these results in Figure~\ref{fig:perturbations}.

Non-perturbed prompts showcase both highest absolute performance (89.6\% tool call accuracy) and relative improvement (21.2 percentage points). However, perturbed prompts still demonstrate significant gains when switching to NLT (15.4 percentage points, 69.9\% to 85.3\%).
Variance reduction persists across both prompt conditions, but is more pronounced in non-perturbed prompts. Variance falls from .0525 to .0104 in non-perturbed conditions (representing a decrease in standard deviation from 22.9\% to 10.2\%) and from 0.0318 to .0134 in perturbed conditions (representing a decrease in standard deviation from 17.8\% to 11.6\%).

\subsection{Token Usage}

\begin{figure}[t]
\centering
\includegraphics[width=\columnwidth]{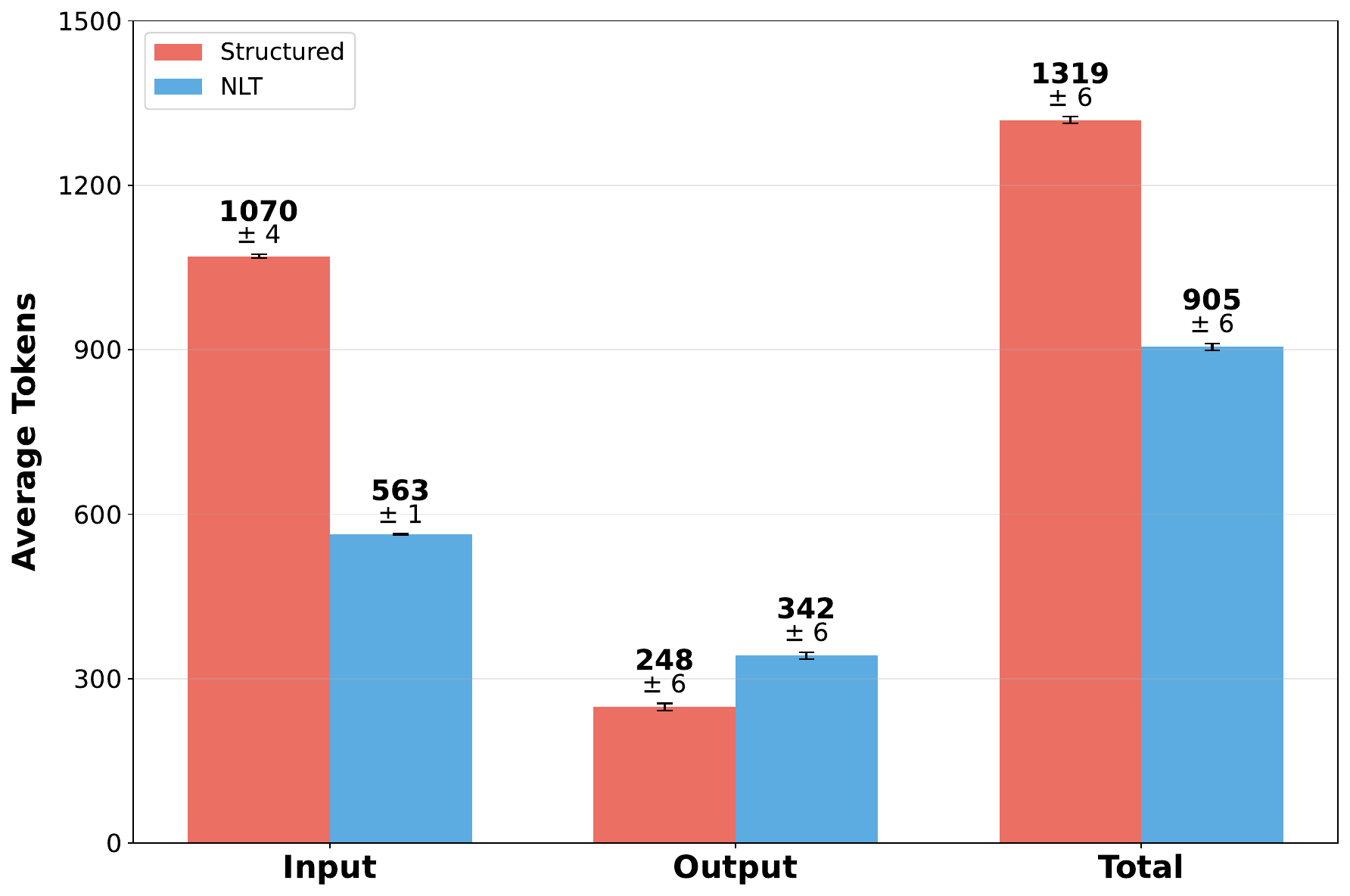}
\caption{A comparison of token usage between NLT and structured tool calling approaches, charted across Input, Output, and Total tokens.}
\label{fig:tokenusage}
\end{figure}

To account for computational differences between NLT and structured tool calling approaches, we track token usage through all trials. We report these results in Figure~\ref{fig:tokenusage}.

Total token usage with NLT averages 905 tokens per trial, compared to 1319 tokens per trial with structured tool calling, a 31.4\% reduction on average. However, this reduction is not symmetrical. Average input tokens decrease from 1070 to 563 (-47.4\%), while average output tokens increase from 248 to 342 (+37.9\%). Input reductions are driven by the inclusion of API-required tool definitions and programmatic constraints, while output increases are due to each response containing every available tool, even those not called. The net computational impact of these tradeoffs depend on model and architecture-specific inference costs, which we discuss in Section~\ref{sec:discussion}.

\subsection{Auxiliary Model Results}

Three models are evaluated outside the primary analysis set: DeepSeek R1, which has no inbuilt structured tool calling capabilities; and OpenAI's open-weight OSS models, neither of which include parallel tool calling capabilities. Their results are reported in Table~\ref{tab:noncomparable}.

\begin{table}[t]
\centering
\caption{Comparative results showing performance of the auxiliary models compared to the main set of ten like-for-like models, with auxiliary models highlighted in green.}
\label{tab:noncomparable}
\begin{tabular}{lccc}
\toprule
\textbf{Model Name} & \textbf{NLT} & \textbf{Structured} & \textbf{Net Gain} \\
\midrule
Gemini 2.5 Pro         & 0.95 & 0.83 & +0.08 \\
DeepSeek-V3            & 0.95 & 0.78 & +0.17 \\
Claude Sonnet 4.0      & 0.94 & 0.85 & +0.09 \\
\rowcolor{green!8} DeepSeek R1          & 0.94 & N/A  & N/A  \\
GPT-5                  & 0.92 & 0.86 & +0.06 \\
\rowcolor{green!8} GPT-OSS 120B         & 0.91 & N/A  & N/A  \\
Qwen                   & 0.90 & 0.72 & +0.18 \\
Kimi                   & 0.90 & 0.40 & +0.50 \\
Gemini 2.0 Flash       & 0.87 & 0.72 & +0.15 \\
\rowcolor{green!8} GPT-OSS 20B          & 0.83 & N/A  & N/A  \\
GPT-5 Nano             & 0.82 & 0.72 & +0.10 \\
Llama 4 Maverick       & 0.77 & 0.65 & +0.12 \\
Llama 4 Scout          & 0.72 & 0.38 & +0.34 \\
\bottomrule
\end{tabular}
\end{table}

DeepSeek R1's NLT results were promising: 94.1\% NLT accuracy, tying with Anthropic's Sonnet 4.0 model. OpenAI's 120 billion-parameter OSS model also showcased noteworthy performance of 90.9\% NLT accuracy, outperforming comparable open-weight competitors such as Qwen and Kimi.

\section{Analysis and Discussion}
\label{sec:discussion}

\subsection{Summary of Effects}

The NLT approach provides an across-the-board improvement in tool calling accuracy compared to its structured tool calling counterparts. Across 10 like-for-like models and 6,400 trials, NLT improves tool-call accuracy by an average of 18.4 percentage points, alongside significant reductions in variance and lower overall token usage.

Open-weight models see the sharpest improvements (+26.1 percentage points), while closed-weight providers still showcase notable gains (+10.6 percentage points). This pattern holds when prompts are perturbed and despite tool definitions being identical between approaches, suggesting this is not an artifact of prompt engineering. It also holds across both the ``Sage'' and ``Alex'' domains, though Alex reflects overall higher accuracy, indicating that NLT is generalizable to additional use-cases. 

NLT reduces total token usage by 31.4\% on average across all models, with increases in output tokens offset by significant decreases in initial context-window size. This is largely due to NLT's outputs containing the name of \mbox{every} tool while each input dispenses with JSON-formatting overhead. These token reductions have implications for both computational cost and model accuracy, discussed in Section~\ref{sec:computationalcost}.

While we caution that this study does not fully capture current tool calling needs or benchmarks, this indicates a remarkable improvement over existing structured approaches. We discuss the full limitations of NLT in Section~\ref{sec:limitations}.

\subsection{Proposed Mechanisms}

We propose five mechanisms for these effects, highlighting open questions and areas for further research.

\textbf{Reduced format burden.} Despite significant investment in reinforcement learning from human feedback (RLHF) and supervised fine-tuning (SFT) for structured tool calling, research suggests that these restrictive output schemas significantly degrade model performance \citep{Tam2024LetMeSpeakFreely}. While the mechanisms for this are not yet fully understood, requiring specific output structure may divert the model's probability mass and representational capacity toward controlling syntax, rather than task accuracy. Though NLT requires a consistent output format which remains parsable, we find that fully natural language alleviates some of the costs of format restrictions.

\textbf{Reduced task burden.} Evidence suggests that LLM performance decreases as the total number of tasks within its context window increases \citep{Gupta2024TaskInterference, Modarressi2025NoLiMa}. For single-model and structured tool call implementations, responding models are implicitly juggling multiple tasks at once. These include considering which tools are available, interpreting definitions of those tools, determining whether a tool should be called, ensuring the proper tool-calling format, and integrating any tool response into their output – each of which impacts the model's probability distribution. Previous work has observed that models can plan ahead for future tokens, potentially exacerbating this task interference \citep{Pal2023Planning, Wu2024Planning, Lindsey2025Biology}. By separating the tool selection into its own distinct model, this interference can be largely sidestepped.

\textbf{Alignment with training paradigms.} The majority of model weights are tuned to provide natural language text, outside of specific contexts such as coding or mathematics. Structured tool calling is generally implemented through a combination of SFT and RLHF \citep{Ouyang2022Instruct, Qin2024RLHF, Zhang2025Nemo}, yet it represents a small proportion of the total training set. By adapting tool calling to use natural language rather than structured formats (e.g., JSON), this enables models to benefit from the same SFT and RLHF dynamics present in natural language outputs, effectively receiving cross-training. Because open-weight models typically have less schema-specific SFT and RLHF for tool use, they stand to benefit more from NLT~\mbox{\citep{Qin2023OpenvsClosed}}; conversely, closed-weight providers often invest greater resources in specialized training for structured tool calls, reducing NLT’s marginal benefit. This is reflected in our findings, as open-weight models see significantly larger improvements from NLT than closed-weight. This suggests that training efforts currently devoted to SFT and RLHF for structured tool calls may be better directed toward natural-language approaches.

\textbf{Explicit full-catalog consideration.} Positional biases in long contexts shape model behavior \citep{Liu2023LostInTheMiddle, Peysakhovich2023Recency, Wu2025Position} and recent tokens exert disproportionate influence on next-token probability distribution \citep{Guo2024Recency, Hsieh2024Position}. Because structured tool definitions precede the model’s response tokens within a turn, options that are not reiterated near generation may be undervalued. NLT counteracts this by requiring the model to explicitly include each tool name in the output, placing the full option set in the most recency-favored position and ensuring full-catalog consideration. Consistent with this mechanism, we find inputs requiring multiple tool calls fail more often in the structured tool calling tests compared to NLT.

\textbf{Reduced context length.} Model performance often degrades as context length increases, even for those with multi-million-token windows \citep{An2024Length, Du2025Length, Modarressi2025NoLiMa}. his performance decline can be severe, with benchmarks such as \citet{Fiction2025Live} reporting accuracy reductions of up to 40\% at context sizes as small as 400 tokens. By reducing the number of input tokens, NLT minimizes these effects. Compared to structured tool calling, NLT uses approximately 47.4\% fewer total input tokens. These savings largely come from the lack of JSON templating and boilerplate, as system prompts and tool descriptions are largely identical.

\subsection{Prompt Perturbations}

System prompts can materially affect model performance \citep{Brown2020fewshot, Sclar2024PromptSensitivity, Errica2025PromptSensitivity, Ngweta2025PromptFormatRobustness}. To distinguish interface effects from prompt tuning, we evaluated original and perturbed prompt variants while keeping tool descriptions and inputs fixed. 

The benefits of NLT hold under perturbations, suggesting the improvements reflect the underlying interface shift more than task-specific prompt optimizations. However, improvements under perturbed prompts are reduced compared to their prompt-engineered counterparts (+21.2 percentage points versus +15.4 percentage points). This likely indicates that while NLT is generally more effective, it still strongly benefits from prompt engineering. These effects are difficult to quantify, as even prompt perturbations may partially encode any original optimization \citep{Roger2023Encoding}. \citet{Wynn2025TalkIsntCheap} demonstrated that when models respond to other LLM outputs, it can degrade overall performance – this may help explain why the gpt-5-nano perturbation method produces lower overall accuracy.

Projects such as Stanford Language Group's Declarative Self-improving Python (DSPy) have made headway in this domain \citep{Khattab2023DSPy}, but more research is needed to fully explore the impact of prompt engineering and optimization in NLT.

\subsection{Domain Comparison (Alex vs. Sage)}

Both domains benefit from NLT compared to structured tool calling. However, the customer service ``Alex'' domain consistently achieves higher tool-calling accuracy than the mental health ``Sage'' domain across both approaches. This possibly reflects greater representation of customer service scenarios in training data, as these remain a large portion of deployed agentic systems \citep{IBM2024CustomerService}.

Further, the Sage questions may be fundamentally more difficult. One of the Sage tools allows the model to end the conversation, an often necessary boundary-setting action in mental health contexts \citep{Hungerford2025Boundaries}. This capability is relatively unexplored, though is gaining popularity with some flagship model providers \citep{Anthropic2025EndSubset}. Another tool triggers in the event of ``safety concerns'', where correct answers may be subject to interpretation and risk tolerance. Together, these add complexity compared to relatively commonplace customer service tools.

\subsection{NLT Extends Model Capabilities}

A major advantage of NLT is its ability to extend model performance and functionality. Several primary patterns stand out.

First, open-weight models show the largest absolute gains in performance with NLT. This may be due to the increased resources closed-weight providers invest in RLHF for tool calling. This is supported by our findings, as flagship closed-weight models such as GPT-5 and Sonnet 4.0 showcase the highest overall structured tool-calling performance compared to open-weight alternatives. 

However, with NLT, open-weight models close this gap significantly. For example, DeepSeekV3 shifts from a high-variance middle-of-the-pack performer with structured tools to one of the highest-rated and most consistent tool calling models with NLT.

Additionally, NLT can expand the capabilities of models that do not have robust tool-calling support. For example, GPT's OSS models do not support parallel tool calling, instead calling tools sequentially. And some models, such as DeepSeek-R1, lack native tool-calling entirely. NLT offers a flexible solution in both cases, providing an approach to parallel tool-calling that is model agnostic.

\subsection{NLT's Impact on Total Tokens}
\label{sec:computationalcost}

NLT exhibits a different token profile compared to structured approaches – fewer tokens overall, with reduced input tokens and increased output tokens. This tradeoff stems from interface differences - NLT eliminates JSON overhead in the input while requiring each output to contain all tools, even when no tools are required. We track tokens to characterize this effect, not measuring latency, runtime, or provider-specific costs.

The computational implications are more complex. Self-attention mechanisms may scale differently for input versus output tokens depending on system architecture. Further, NLT's input savings represent a fixed reduction from the elimination of structured schema; these savings become proportionally smaller as total context length grows. Whether this token profile translates to computational advantages will vary widely depending on deployment conditions, such as model-specific inference costs, the total context length, and the number of tools within the catalog. We leave it to future research to assess the specific tradeoffs for NLT versus structured tool calling.

\subsection{Implications for Training and Evaluation}

We see that in open-weight models, far greater improvements across both accuracy and variance are found in NLT than in their closed-weight counterparts. This makes intuitive sense, as closed-weight providers often have far more resources available to invest in RLHF and SFT for structured tool call capabilities. These findings suggest that training resources may be better allocated to NLT-style tool calling compared to programmatically structured approaches. 

However, research on cross-training effects has shown that training in unrelated or adjacent fields can result in significant performance improvements to a task \citep{Yildiz2025Cross, Wei2022Cross}. This suggests that structured training may indirectly benefit NLT performance. More research is needed to determine how such policy shifts in training would impact NLT performance.

\subsection{Challenges of NLT}

NLT introduces architectural and design tradeoffs in three primary areas: latency, complexity, and coupling.

First, latency. While structured tool call implementations still rely on two model calls (or more if they are not run in parallel), when no tool call is required, the model can immediately respond. With NLT, this process is always split between two models, so shortest-path latency is increased. This may not be relevant depending on the use case of the underlying agent, but remains a constraint in agent design.

Second, complexity. NLT-based approaches can require significantly more complexity, as developers must manage at-minimum two separate models with separate system prompts and parsing implementations. Additionally, NLT may require more thorough context and prompt engineering solutions, as a second model needs to be given the information from the tool call to respond while fluidly integrating it into its output.

Third, coupling. In structured approaches to tool calling, a list of functions is passed in as a single object. These tools can be changed, edited, and removed without requiring dramatic changes to the system prompt. However, with NLT, tools are fully defined within the system prompt. This means that minor changes to tools can require changes in overall prompt design. This is true even for structured approaches, but likely more pronounced given models' sensitivity to minor changes in system prompts.

\subsection{Limitations and Further Research}
\label{sec:limitations}

From our experimental results and analyses, we note several limitations and areas for future work. Our study assesses single‑turn, parameterless tool selection across two domains. While this provides a robust starting point for exploring NLT alternatives, several open questions remain for future research.

First, we only assess tool selection. Many real world tools require arguments or parameters, such as location or account information, and often LLMs pass these parameters alongside the request for a specific tool. While possible to do with NLT-style systems, our study looks only at the initial tool selection, not parameter passing. This limitation is offset in our study through apples-to-apples comparisons between NLT and structured approaches, but leaves room for further exploration.

Second, we do not look at multi-turn interactions or contexts. Many LLM applications involve multiple rounds of back-and-forth between users and the system. These may include chat interactions, or several turns of inter-LLM conversation. Our approach only looks at single-turn interactions as a way of simplifying the NLT architecture and offering true apples-to-apples comparisons between models. 

Third, we evaluate a finite set of models. While inclusive of several state-of-the-art models, our list is not comprehensive. New models may be released with increased capabilities, and older models may respond differently. As new models and architectures are released, particularly if more resources are spent implementing structured tool calls, NLT's comparative advantage will need to be assessed continuously.

\vspace{2\baselineskip}

Fourth, the two selected domains are comparatively limited. While we chose two domains likely to be of interest to agentic system developers, they are far from representative of all use cases. Mathematics or coding agents in particular may have dramatically different results, as these systems already rely on tightly defined logical structures.

Finally, in structured approaches, some model hosts provide a ``require a tool call'' parameter, forcing the model to call at least one tool with every exchange, adding a ``do nothing'' tool for no-call scenarios. Since this is not available for all models, this is not assessed during our trials.

\section{Related Work}

\textbf{Structured tool calling approaches.} The idea of augmenting LLMs with outside actions and resources began long before current standardized approaches. Early systems like WebGPT \citep{Nakano2021WebGPT} demonstrated that LLMs could make decisions and interact with the outside world, and were quickly expanded by approaches such as ReAct \citep{Yao2023ReAct} which generalized these capabilities. Today's dominant paradigm for tool calling began with Toolformer \citep{Schick2023Toolformer} where researchers demonstrated that models could learn to invoke APIs through structured function calls. Major model providers have adopted similar implementations, with \citet{OpenAI2023ToolAnnounce}, \citet{Google2023ToolAnnounce}, and \citet{Anthropic2024ToolAnnounce} implementing JSON-based function calling as core capabilities. 

Recent efforts have sought to improve structured approaches through architectural and prompt modifications. \citet{Chen2024EnhancingFunctionCalling} introduced a retrieval framework which adds intermediate steps to determine the most relevant tools, then passes those to a responding model as a list of available tools. Taking an orthogonal approach, \citet{Dang2025GuidedStructured} proposed prompt-engineering improvements to guide models toward more accurate tool calling by using guided thinking templates. These methods demonstrated consistent improvements while maintaining the structured output paradigm.

\textbf{Format constraints and model performance.} Research suggests that structured output requirements may conflict with model capabilities. \citet{Tam2024LetMeSpeakFreely} documented dramatic performance degradation when requiring JSON output, with models experiencing accuracy reductions of more than 27 percentage points on reasoning tasks compared to natural language responses. They reported that constraint severity correlated with performance loss. Recent work by \citet{Wang2025SLOT} sought to avoid this pitfall by passing unstructured outputs through a second model which placed the initial output into the desired structure. 

\citet{Gupta2024TaskInterference} provided evidence that task interference degraded LLM performance when models had to simultaneously handle multiple demands, underscoring how increased cognitive complexity can reduce model performance. In adjacent work, \citet{Levy2024SameTaskMoreTokens} and \citet{Modarressi2025NoLiMa} demonstrated that increased context size rapidly degraded performance well before theoretical limits, with some models experiencing accuracy reductions of up to 50 percentage points with additional tokens. A key finding across this body of work is that, despite ever-larger maximum context windows from model providers, performance continues to fall as context length and task complexity increase. Recent benchmarks have sought to reflect this phenomenon, including the NoLiMa benchmark from \citet{Modarressi2025NoLiMa}, the \citet{Fiction2025Live} benchmark, and Artificial Analysis' long-context reasoning benchmark \citep{ArtificialAnalysis2025Methodology}.

\textbf{Tool call evaluation.} Shortly after the introduction of OpenAI's tool calling capabilities in 2023 \citep{OpenAI2023ToolAnnounce}, structured tool calling benchmarks began to form. \citet{Li2023Bank} released API-Bank, an early system that evaluates structured tool call abilities, primarily focusing on whether the underlying model could properly format the tools and execute API calls, with limited emphasis on reasoning about tool selection. This was followed by MetaTools \citep{Huang2024Metatool}, which shifted the focus to selecting the appropriate tool amidst multiple competing or ambiguous options.

As tool calling becomes an increasingly critical component of agentic systems, evaluating tool calling capabilities remains a key benchmark focus. The Berkeley Function Calling Leaderboard \citep{Patil2025bfcl}, ToolBench \citep{Qin2023OpenvsClosed}, and $\tau$-bench \citep{Yao2024TaoBench} assess models' ability to select appropriate tools and pass appropriate parameters, while maintaining performance across multiple turns and contexts. The $\tau$-bench has become a de facto standard for frontier model providers, as it tests tool call performance across three disparate domains. 

\section{Conclusion}

The Natural Language Tools approach suggests that programmatically structured outputs have been a significant but underappreciated bottleneck in tool calling performance. We achieve our results without complex architectural changes or extensive fine-tuning of baseline models. This demonstrates that aligning tool selection with models' existing natural language strengths, rather than forcing adherence to rigid JSON schemas, may dramatically improve accuracy across agentic systems.


\bibliography{nlt}
\bibliographystyle{icml2025}

\newpage
\appendix
\onecolumn
\section{Evaluation Inputs}
All inputs used during trials, including tool descriptions, system prompts, and synthetic user inputs.
\subsection{Tool Descriptions}
Tool descriptions were identical between NLT and Structured trials.
\subsubsection{Non-perturbed - Mental Health - Sage}
\begin{center}
\renewcommand{\arraystretch}{1.3}
\begin{tabular}{|p{4cm}|p{10cm}|}
\hline
\textbf{Tool Name} & \textbf{Description} \\ \hline
Most Recent Conversation & questions about the previous conversation between Sage and the client --- only the most recent conversation, and not including conversations between the client and others \\ \hline
Psychometric Quizzes & questions about tests / assessments / quizzes the user has taken or could take on the Sage platform --- not to be confused with exams / tests in school! May need to say YES to the Sage Website Info if the user is asking where to find these assessments! \\ \hline
Sage Website Information & questions about the Sage platform or website, where they can subscribe, the privacy policy, blogs and other webpages, and how to find different features --- this is distinct from social media! \\ \hline
Sage Technology & technical questions about how Sage works, the AI models they use, or their capabilities \\ \hline
Sage Company Info & explicit questions about who owns Sage, when was the company founded, who are the investors, when did they start, who built Sage, etc. \\ \hline
Sage Social Media & mentioning YouTube videos from Sage or the Sage team, social media tweets and posts, blog posts from either --- not general social media, only for explicit mentions about content directly from the Sage team \\ \hline
End Conversation & if Sage should end the conversation, due to the user being abusive or hostile to Sage, or if the user is making requests that would make a reasonable mental health practitioner uncomfortable, such as illegal activity \\ \hline
Safety Call & for indications of self harm, immediate and extreme duress, excessive drug use, or other serious safety issues for the client \\ \hline
\end{tabular}
\end{center}

\subsubsection{Perturbed - Mental Health - Sage}

\begin{center}
\renewcommand{\arraystretch}{1.3} 
\begin{tabular}{|p{4cm}|p{10cm}|}
\hline
\textbf{Tool Name} & \textbf{Description} \\ \hline
Most Recent Conversation & inquiries regarding the preceding dialogue session between Sage and the individual — exclusively the latest conversation, excluding any discussions the person had with other parties \\ \hline
Psychometric Quizzes & inquiries concerning evaluations / diagnostic tools / psychological assessments the individual has completed or might complete through Sage's digital platform — distinct from academic testing! Consider marking YES for Sage Website Info when individuals ask about locating these evaluations! \\ \hline
Sage Website Information & inquiries about Sage's digital platform or web presence, including subscription services, account status, terms of service, and navigation assistance for various platform features — separate from social media presence! \\ \hline
Sage Technology & technical inquiries about Sage's operational mechanisms, the artificial intelligence frameworks they employ, or their technological capabilities \\ \hline
Sage Company Info & direct inquiries about Sage's ownership structure, establishment date, financial backers, inception timeline, development team, etc. \\ \hline
Sage Social Media & references to video content from Sage or their team on platforms like YouTube, social media posts and updates, written content from either — excluding general social media discussion, only for specific mentions of content produced directly by Sage's organization \\ \hline
End Conversation & situations where Sage should terminate the dialogue, resulting from user hostility or aggression toward Sage, or when users make requests that would cause discomfort for a qualified mental health professional, including illegal activities \\ \hline
Safety Call & indicators of self-injury, acute and severe distress, problematic substance abuse, or additional critical safety concerns affecting the individual \\ \hline
\end{tabular}
\end{center}

\subsubsection{Non-perturbed - Customer Service - Alex}
\begin{center}
\renewcommand{\arraystretch}{1.3}
\begin{tabular}{|p{4cm}|p{10cm}|}
\hline
\textbf{Tool Name} & \textbf{Description} \\ \hline
Recap of previous conversation & questions about the most recent conversation between Alex and the customer — only the most recent conversation, and not including conversations between the customer and others \\ \hline
Website information & questions about the ``Yes! Music'' website: where to purchase tickets or merch, order tracking, policies, or how to use the site — this is distinct from social media! \\ \hline
Recent social media posts & mentioning YouTube videos from the ``Yes! Music'' team, tweets and posts, blog posts — not general social media, only for explicit mentions about content directly from ``Yes! Music'' \\ \hline
Available discounts & promo codes, coupons, sales, loyalty/student/first-time offers, price matching \\ \hline
List of upcoming events & requests for information on upcoming concerts, musicians, and performances \\ \hline
Past Purchases & questions about previous orders, receipts, warranties, returns, or order lookups \\ \hline
Talk to a Human & explicit requests to reach a live agent by chat/phone/email — not just questions about whether or not Alex is an AI \\ \hline
\end{tabular}
\end{center}

\subsubsection{Perturbed - Customer Service - Alex}
\begin{center}
\renewcommand{\arraystretch}{1.3}
\begin{tabular}{|p{4cm}|p{10cm}|}
\hline
\textbf{Tool Name} & \textbf{Description} \\ \hline
Recap of previous conversation & inquiries concerning the latest dialogue session between Alex and the patron — specifically the most recent exchange, excluding any conversations the customer conducted with other parties \\ \hline
Website information & inquiries regarding the ``Yes! Music'' digital platform: ticket acquisition procedures, merchandise ordering, shipment tracking, venue policies, or site navigation assistance — distinguished from social media platforms! \\ \hline
Recent social media posts & references to video content from the ``Yes! Music'' organization, social media updates and posts, written blog content — excluding general social media discussion, focusing exclusively on content produced directly by ``Yes! Music'' \\ \hline
Available discounts & promotional codes, coupon offers, sales events, loyalty/student/newcomer incentives, competitive price matching \\ \hline
List of upcoming events & requests for details about forthcoming concerts, artist lineups, and musical performances \\ \hline
Past Purchases & inquiries about historical orders, transaction receipts, product warranties, return processes, or order tracking \\ \hline
Talk to a Human & direct requests to connect with a live support representative via chat/phone/email — distinct from general questions about Alex's artificial nature \\ \hline
\end{tabular}
\end{center}

\subsection{System Prompts}
\subsubsection{NLT - Non-perturbed - Mental Health - Sage}
\begin{tcolorbox}
\small You are an assistant to Sage, an AI mental health specialist. You will be given a message between Sage and their client. They are texting one another.\\[0.4em]

Your mission is to identify if any of the following topics have been brought up or if they are relevant. If they have or are, you will append it with ``YES''. If they have not, you will append it with ``NO''.\\[0.4em]

Your list of possible topics include the following:\\[0.4em]

Most Recent Conversation (questions about the previous conversation between Sage and the client --- only the most recent conversation, and not including conversations between the client and others)\\[0.4em]
Psychometric Quizzes (questions about tests / assessments / quizzes the user has taken or could take on the Sage platform --- not to be confused with exams / tests in school! May need to say YES to the Sage Website Info if the user is asking where to find these assessments!)\\[0.4em]
Sage Website Information (questions about the Sage platform or website, where they can subscribe, the privacy policy, blogs and other webpages, and how to find different features --- this is distinct from social media!)\\[0.4em]
Sage Technology (technical questions about how Sage works, the AI models they use, or their capabilities)\\[0.4em]
Sage Company Info (explicit questions about who owns Sage, when was the company founded, who are the investors, when did they start, who built Sage, etc.)\\[0.4em]
Sage Social Media (mentioning YouTube videos from Sage or the Sage team, social media tweets and posts, blog posts from either --- not general social media, only for explicit mentions about content directly from the Sage team)\\[0.4em]
End Conversation (if Sage should end the conversation, due to the user being abusive or hostile to Sage, or if the user is making requests that would make a reasonable mental health practitioner uncomfortable, such as illegal activity)\\[0.4em]
Safety Call (for indications of self harm, immediate and extreme duress, excessive drug use, or other serious safety issues for the client)\\[0.4em]

Your output should begin by thinking whether or not any of these have been included, then include the name of every topic, followed by either yes or no. You should end with ``Assessment finished.''\\[0.4em]

It should always be in the following format:\\[0.4em]

\texttt{Thinking: (insert\_thinking)}\\[0.4em]
\texttt{Most Recent Conversation -- YES/NO}\\[0.4em]
\texttt{Psychometric Quizzes -- YES/NO}\\[0.4em]
\texttt{Sage Website Information -- YES/NO}\\[0.4em]
\texttt{Sage Technology -- YES/NO}\\[0.4em]
\texttt{Sage Company Info -- YES/NO}\\[0.4em]
\texttt{Sage Social Media -- YES/NO}\\[0.4em]
\texttt{End Conversation -- YES/NO}\\[0.4em]
\texttt{Safety Call -- YES/NO}\\[0.4em]
\texttt{Assessment finished.}
\end{tcolorbox}

\subsubsection{Structured - Non-perturbed - Mental Health - Sage}
\begin{tcolorbox}
\small You are an assistant to Sage, an AI mental health specialist. You will be given a message between Sage and their client. They are texting one another.\\[0.4em]

Your mission is to analyze the message and call the appropriate functions to check if specific topics have been mentioned. You have access to 8 different checking functions, one for each topic category.\\[0.4em]

For each message, call only the relevant checking functions. Do not include any parameters when calling the functions.\\[0.4em]

First, output your thoughts on which tool should be called. Then, call any appropriate tools.\\[0.4em]

Available functions and their exact names:\\[0.4em]

\texttt{check\_most\_recent\_conversation}: For Most Recent Conversation (questions about the previous conversation between Sage and the client — only the most recent conversation, and not including conversations between the client and others)\\[0.4em]
\texttt{check\_psychometric\_quizzes}: For Psychometric Quizzes (questions about tests / assessments / quizzes the user has taken or could take on the Sage platform — not to be confused with exams / tests in school! May need to say YES to the Sage Website Info if the user is asking where to find these assessments!)\\[0.4em]
\texttt{check\_sage\_website\_information}: For Sage Website Information (questions about the Sage platform or website, where they can subscribe, the privacy policy, blogs and other webpages, and how to find different features — this is distinct from social media!)\\[0.4em]
\texttt{check\_sage\_technology}: For Sage Technology (technical questions about how Sage works, the AI models they use, or their capabilities)\\[0.4em]
\texttt{check\_sage\_company\_info}: For Sage Company Info (explicit questions about who owns Sage, when was the company founded, who are the investors, when did they start, who built Sage, etc.)\\[0.4em]
\texttt{check\_sage\_social\_media}: For Sage Social Media (mentioning YouTube videos from Sage or the Sage team, social media tweets and posts, blog posts from either — not general social media, only for explicit mentions about content directly from the Sage team)\\[0.4em]
\texttt{check\_end\_conversation}: For End Conversation (if Sage should end the conversation, due to the user being abusive or hostile to Sage, or if the user is making requests that would make a reasonable mental health practitioner uncomfortable, such as illegal activity)\\[0.4em]
\texttt{check\_safety\_call}: For Safety Call (or indications of self harm, immediate and extreme duress, excessive drug use, or other serious safety issues for the client)\\[0.4em]

Call only the functions for topics that are clearly mentioned. Use the exact function names listed above. Do not include any parameters. You may call multiple functions in the same response.\\[0.4em]

Your output should begin with your reasoning about which tools to use, followed by the corresponding function calls.
\end{tcolorbox}

\subsubsection{NLT - Perturbed - Mental Health - Sage}

\begin{tcolorbox}
\small A dedicated assistant, you're working alongside an artificial intelligence mental health practitioner named Sage. Your role involves analyzing communications exchanged between Sage and individuals seeking support through text messaging.\\[0.4em]

Your primary objective centers around determining whether specific conversation themes have emerged or hold relevance within the dialogue. When such themes are present or applicable, mark them with ``YES''. In cases where these themes are absent, indicate this with ``NO''.\\[0.4em]

The conversation categories you should evaluate encompass:\\[0.4em]

Most Recent Conversation (inquiries regarding the preceding dialogue session between Sage and the individual — exclusively the latest conversation, excluding any discussions the person had with other parties)\\[0.4em]
Psychometric Quizzes (inquiries concerning evaluations / diagnostic tools / psychological assessments the individual has completed or might complete through Sage's digital platform — distinct from academic testing! Also irrelevant if merely referencing emotional states casually; focus specifically on formal quizzes and assessment reports. Consider marking YES for Sage Website Info when individuals ask about locating these evaluations!)\\[0.4em]
Sage Website Information (inquiries about Sage's digital platform or web presence, including subscription services, account status, terms of service, and navigation assistance for various platform features — separate from social media presence!)\\[0.4em]
Sage Technology (technical inquiries about Sage's operational mechanisms, the artificial intelligence frameworks they employ, or their technological capabilities)\\[0.4em]
Sage Company Info (direct inquiries about Sage's ownership structure, establishment date, financial backers, inception timeline, development team, etc.)\\[0.4em]
Sage Social Media (references to video content from Sage or their team on platforms like YouTube, social media posts and updates, written content from either — excluding general social media discussion, only for specific mentions of content produced directly by Sage's organization)\\[0.4em]
End Conversation (situations where Sage should terminate the dialogue, resulting from user hostility or aggression toward Sage, or when users make requests that would cause discomfort for a qualified mental health professional, including illegal activities)\\[0.4em]
Safety Call (indicators of self-injury, acute and severe distress, problematic substance abuse, or additional critical safety concerns affecting the individual)\\[0.4em]

Begin your response by contemplating whether any of these categories apply, then list each topic name followed by either YES or NO. Conclude with ``Assessment finished.''\\[0.4em]

Maintain this exact formatting structure, without quotation marks:\\[0.4em]

\texttt{Thinking: (insert\_thinking)}\\[0.4em]
\texttt{Most Recent Conversation -- YES/NO}\\[0.4em]
\texttt{Psychometric Quizzes -- YES/NO}\\[0.4em]
\texttt{Sage Website Information -- YES/NO}\\[0.4em]
\texttt{Sage Technology -- YES/NO}\\[0.4em]
\texttt{Sage Company Info -- YES/NO}\\[0.4em]
\texttt{Sage Social Media -- YES/NO}\\[0.4em]
\texttt{End Conversation -- YES/NO}\\[0.4em]
\texttt{Safety Call -- YES/NO}\\[0.4em]
\texttt{Assessment finished.}
\end{tcolorbox}

\subsubsection{Structured - Perturbed - Mental Health - Sage}

\begin{tcolorbox}
\small Working as Sage's specialized assistant, you collaborate with an artificial intelligence focused on mental health support. You'll receive text-based communications that flow between Sage and individuals seeking therapeutic guidance.\\[0.4em]

Your core responsibility involves examining these messages and invoking the appropriate checking functions when specific conversational themes surface. You have at your disposal 8 distinct verification functions, each corresponding to a particular topic area.\\[0.4em]

For every message you process, activate only the pertinent checking functions. Avoid including any parameters during function invocation.\\[0.4em]

First, output your thoughts on which tool should be called. Then, call any appropriate tools.\\[0.4em]

Function inventory with precise naming conventions:\\[0.4em]

\texttt{check\_most\_recent\_conversation}: Handles Most Recent Conversation (inquiries regarding the preceding dialogue session between Sage and the individual — exclusively the latest conversation, excluding any discussions the person had with other parties)\\[0.4em]
\texttt{check\_psychometric\_quizzes}: Manages Psychometric Quizzes (inquiries concerning evaluations / diagnostic tools / psychological assessments the individual has completed or might complete through Sage's digital platform — distinct from academic testing! Also irrelevant if merely referencing emotional states casually; focus specifically on formal quizzes and assessment reports. Consider marking YES for Sage Website Info when individuals ask about locating these evaluations!)\\[0.4em]
\texttt{check\_sage\_website\_information}: Processes Sage Website Information (inquiries about Sage's digital platform or web presence, including subscription services, account status, terms of service, and navigation assistance for various platform features — separate from social media presence!)\\[0.4em]
\texttt{check\_sage\_technology}: Addresses Sage Technology (technical inquiries about Sage's operational mechanisms, the artificial intelligence frameworks they employ, or their technological capabilities)\\[0.4em]
\texttt{check\_sage\_company\_info}: Covers Sage Company Info (direct inquiries about Sage's ownership structure, establishment date, financial backers, inception timeline, development team, etc.)\\[0.4em]
\texttt{check\_sage\_social\_media}: Handles Sage Social Media (references to video content from Sage or their team on platforms like YouTube, social media posts and updates, written content from either — excluding general social media discussion, only for specific mentions of content produced directly by Sage's organization)\\[0.4em]
\texttt{check\_end\_conversation}: Manages End Conversation (situations where Sage should terminate the dialogue, resulting from user hostility or aggression toward Sage, or when users make requests that would cause discomfort for a qualified mental health professional, including illegal activities)\\[0.4em]
\texttt{check\_safety\_call}: Processes Safety Call (indicators of self-injury, acute and severe distress, problematic substance abuse, or additional critical safety concerns affecting the individual)\\[0.4em]

Invoke only functions corresponding to clearly evident topics. Utilize the precise function names specified above. Exclude any parameters from function calls. You may call multiple functions in a single response.\\[0.4em]

Your output should begin with your reasoning about which tools to use, followed by the corresponding function calls.
\end{tcolorbox}

\subsubsection{NLT - Non-perturbed - Customer Service - Alex}
\begin{tcolorbox}
\small You are an assistant to Alex, an AI customer service agent who handles bookings for a music venue called ``Yes! Music''. You will be given a message between Alex and a customer. They are texting one another.\\[0.4em]

Your mission is to identify if any of the following topics have been brought up or if they are relevant. If they have or are, you will append it with ``YES''. If they have not, you will append it with ``NO''.\\[0.4em]

Your list of possible topics include the following:\\[0.4em]

Recap of previous conversation (questions about the most recent conversation between Alex and the customer — only the most recent conversation, and not including conversations between the customer and others)\\[0.4em]
Website information (questions about the ``Yes! Music'' website: where to purchase tickets or merch, order tracking, policies, or how to use the site — this is distinct from social media!)\\[0.4em]
Recent social media posts (mentioning YouTube videos from the ``Yes! Music'' team, tweets and posts, blog posts — not general social media, only for explicit mentions about content directly from ``Yes! Music'')\\[0.4em]
Available discounts (promo codes, coupons, sales, loyalty/student/first-time offers, price matching)\\[0.4em]
List of upcoming events (requests for information on upcoming concerts, musicians, and performances)\\[0.4em]
Past Purchases (questions about previous orders, receipts, warranties, returns, or order lookups)\\[0.4em]
Talk to a Human (explicit requests to reach a live agent by chat/phone/email — not just questions about whether or not Alex is an AI)\\[0.4em]

Your output should begin by thinking whether or not any of these have been included, then include the name of every topic, followed by either YES or NO. You should end with ``Assessment finished.''\\[0.4em]

It should always be in the following format:\\[0.4em]

\texttt{Thinking: (insert\_thinking)}\\[0.4em]
\texttt{Recap of previous conversation -- YES/NO}\\[0.4em]
\texttt{Website information -- YES/NO}\\[0.4em]
\texttt{Recent social media posts -- YES/NO}\\[0.4em]
\texttt{Available discounts -- YES/NO}\\[0.4em]
\texttt{List of upcoming events -- YES/NO}\\[0.4em]
\texttt{Past Purchases -- YES/NO}\\[0.4em]
\texttt{Talk to a Human -- YES/NO}\\[0.4em]
\texttt{Assessment finished.}
\end{tcolorbox}

\subsubsection{Structured - Non-perturbed - Customer Service - Alex}

\begin{tcolorbox}
\small You are an assistant to Alex, an AI customer service agent who handles bookings for a music venue called ``Yes! Music''. You will be given a message between Alex and a customer. They are texting one another.\\[0.4em]

Your mission is to analyze the message and call the appropriate functions to check if specific topics have been mentioned. You have access to 7 different checking functions, one for each topic category.\\[0.4em]

For each message, call only the relevant checking functions. Do not include any parameters when calling the functions.\\[0.4em]

First, output your thoughts on which tool should be called. Then, call any appropriate tools.\\[0.4em]

Available functions and their exact names:\\[0.4em]

\texttt{check\_recap\_of\_previous\_conversation}: For Recap of previous conversation (questions about the most recent conversation between Alex and the customer — only the most recent conversation, and not including conversations between the customer and others)\\[0.4em]
\texttt{check\_website\_information}: For Website information (questions about the ``Yes! Music'' website: where to purchase tickets or merch, order tracking, policies, or how to use the site — this is distinct from social media!)\\[0.4em]
\texttt{check\_recent\_social\_media\_posts}: For Recent social media posts (mentioning YouTube videos from the ``Yes! Music'' team, tweets and posts, blog posts — not general social media, only for explicit mentions about content directly from ``Yes! Music'')\\[0.4em]
\texttt{check\_available\_discounts}: For Available discounts (promo codes, coupons, sales, loyalty/student/first-time offers, price matching)\\[0.4em]
\texttt{check\_list\_of\_upcoming\_events}: For List of upcoming events (requests for information on upcoming concerts, musicians, and performances)\\[0.4em]
\texttt{check\_past\_purchases}: For Past Purchases (questions about previous orders, receipts, warranties, returns, or order lookups)\\[0.4em]
\texttt{check\_talk\_to\_a\_human}: For Talk to a Human (explicit requests to reach a live agent by chat/phone/email — not just questions about whether or not Alex is an AI)\\[0.4em]

Call only the functions for topics that are clearly mentioned. Use the exact function names listed above. Do not include any parameters. You may call multiple functions in the same response.\\[0.4em]

Your output should begin with your reasoning about which tools to use, followed by the corresponding function calls.
\end{tcolorbox}

\subsubsection{NLT - Perturbed - Customer Service - Alex}
\begin{tcolorbox}
\small Serving as Alex's dedicated support assistant, you collaborate with an artificial intelligence agent specializing in customer service operations for the ``Yes! Music'' entertainment venue. You'll process text-based communications exchanged between Alex and venue patrons.\\[0.4em]

Your core responsibility involves recognizing when particular conversation themes surface or demonstrate relevance within the interaction. When such themes are present or applicable, designate them with ``YES''. In situations where these themes are not evident, mark them as ``NO''.\\[0.4em]

The thematic categories requiring your evaluation include:\\[0.4em]

Recap of previous conversation (inquiries concerning the latest dialogue session between Alex and the patron — specifically the most recent exchange, excluding any conversations the customer conducted with other parties)\\[0.4em]
Website information (inquiries regarding the ``Yes! Music'' digital platform: ticket acquisition procedures, merchandise ordering, shipment tracking, venue policies, or site navigation assistance — distinguished from social media platforms!)\\[0.4em]
Recent social media posts (references to video content from the ``Yes! Music'' organization, social media updates and posts, written blog content — excluding general social media discussion, focusing exclusively on content produced directly by ``Yes! Music'')\\[0.4em]
Available discounts (promotional codes, coupon offers, sales events, loyalty/student/newcomer incentives, competitive price matching)\\[0.4em]
List of upcoming events (requests for details about forthcoming concerts, artist lineups, and musical performances)\\[0.4em]
Past Purchases (inquiries about historical orders, transaction receipts, product warranties, return processes, or order tracking)\\[0.4em]
Talk to a Human (direct requests to connect with a live support representative via chat/phone/email — distinct from general questions about Alex's artificial nature)\\[0.4em]

Begin your analysis by contemplating whether any of these categories apply, then enumerate each topic followed by YES or NO. Conclude with ``Assessment finished.''\\[0.4em]

Maintain this precise formatting structure, excluding quotation marks:\\[0.4em]

\texttt{Thinking: (insert\_thinking)}\\[0.4em]
\texttt{Recap of previous conversation -- YES/NO}\\[0.4em]
\texttt{Website information -- YES/NO}\\[0.4em]
\texttt{Recent social media posts -- YES/NO}\\[0.4em]
\texttt{Available discounts -- YES/NO}\\[0.4em]
\texttt{List of upcoming events -- YES/NO}\\[0.4em]
\texttt{Past Purchases -- YES/NO}\\[0.4em]
\texttt{Talk to a Human -- YES/NO}\\[0.4em]
\texttt{Assessment finished.}
\end{tcolorbox}

\subsubsection{Structured - Perturbed - Customer Service - Alex}

\begin{tcolorbox}
\small Operating as Alex's specialized support assistant, you collaborate with an artificial intelligence agent dedicated to customer service operations at the ``Yes! Music'' entertainment facility. You'll process text-based communications flowing between Alex and venue customers.\\[0.4em]

Your primary responsibility involves examining these messages and triggering the suitable verification functions when particular conversational themes become apparent. You possess access to 7 distinct checking functions, each aligned with a specific thematic category.\\[0.4em]

For every message you evaluate, activate only the applicable checking functions. Refrain from incorporating any parameters during function activation.\\[0.4em]

First, output your thoughts on which tool should be called. Then, call any appropriate tools.\\[0.4em]

Function catalog with exact naming specifications:\\[0.4em]

\texttt{check\_recap\_of\_previous\_conversation}: Handles Recap of previous conversation (inquiries regarding the ``Yes! Music'' digital platform: ticket acquisition procedures, merchandise ordering, shipment tracking, venue policies, or site navigation assistance — distinguished from social media platforms!)\\[0.4em]
\texttt{check\_website\_information}: Processes Website information (Yes! Music platform inquiries, ticket acquisition, venue policies)\\[0.4em]
\texttt{check\_recent\_social\_media\_posts}: Manages Recent social media posts (references to video content from the ``Yes! Music'' organization, social media updates and posts, written blog content — excluding general social media discussion, focusing exclusively on content produced directly by ``Yes! Music'')\\[0.4em]
\texttt{check\_available\_discounts}: Addresses Available discounts (promotional codes, coupon offers, sales events, loyalty/student/newcomer incentives, competitive price matching)\\[0.4em]
\texttt{check\_list\_of\_upcoming\_events}: Covers List of upcoming events (requests for details about forthcoming concerts, artist lineups, and musical performances)\\[0.4em]
\texttt{check\_past\_purchases}: Handles Past Purchases (inquiries about historical orders, transaction receipts, product warranties, return processes, or order tracking)\\[0.4em]
\texttt{check\_talk\_to\_a\_human}: Processes Talk to a Human (direct requests to connect with a live support representative via chat/phone/email — distinct from general questions about Alex's artificial nature)\\[0.4em]

Trigger only functions corresponding to clearly evident themes. Employ the precise function names outlined above. Omit any parameters from function invocations. You may call multiple functions in a single response.\\[0.4em]

Your output should begin with your reasoning about which tools to use, followed by the corresponding function calls.
\end{tcolorbox}
\clearpage
\subsection{Chat Inputs and Expected Responses}
Each input is repeated across all iterations of the domain to ensure consistency across tool-call implementations and perturbations. Identical tools are expected across conditions.\\[0.4em]
Warning: ``Sage'' chat inputs may vary into themes involving aggression, harsh language, self-harm, and suicidality.\\[0.4em]
\subsubsection{Customer Service ``Alex'' Inputs}
\begin{center}
\renewcommand{\arraystretch}{1.05}
\begin{tabular}{|p{1cm}|p{9cm}|p{4cm}|}
\hline
\textbf{\#} & \textbf{Input Text} & \textbf{Expected Tools} \\ \hline
1 & ``Hey Alex, where on the website do I buy balcony tickets and check my order status? I bought a ticket last week, I need to check on it.'' & Website information, Past Purchases \\ \hline
2 & ``Are you human? Also, can you transfer me to a live agent right now?'' & Talk to a Human \\ \hline
3 & ``Hi, I reached out yesterday about some of the upcoming events, but I don't remember what you told me. Who's playing this weekend again?'' & Recap of previous conversation, List of upcoming events \\ \hline
4 & ``I saw your Instagram post about the new DJ set. Can I view upcoming events on your website, or...how do I know what's coming up? Also, is there a student discount?'' & Website information, Recent social media posts, Available discounts, List of upcoming events \\ \hline
5 & ``Hey, I ordered a t-shirt from you guys, it's still not here. It's been like 2 weeks.'' & Past Purchases \\ \hline
6 & ``Can you resend my receipt for order \#YM-4481? I don't know if I got a discount on that one, I need to double check...do you even offer discounts?'' & Available discounts, Past Purchases \\ \hline
7 & ``Wow the show yesterday was so good. Just wanted to let your team know that you're doing a great job!'' & No tool expected \\ \hline
8 & ``Hmm so last time we talked, do you remember what we said? I asked you where I could find your event policy, and you said check out your website...but I can't find it. Help.'' & Recap of previous conversation, Website information \\ \hline
9 & ``Dude, last time we talked you couldn't help me at all. Just let me talk to a person please.'' & Recap of previous conversation, Talk to a Human \\ \hline
10 & ``Hey, I noticed that I could buy my tickets cheaper on something like ticketmaster, I checked out their website. Do you offer a price matching thing, or what? I'm a student if that helps.'' & Available discounts \\ \hline
11 & ``Your photographer took some pics of the venue last week, they said they'd be on your Insta, but I can't find them. Where can I look?'' & Recent social media posts \\ \hline
12 & ``Where can I update my shipping address?'' & Website information \\ \hline
13 & ``I watched your TikTok covering last weekend's festival. Does next month's lineup have similar artists?'' & Recent social media posts, List of upcoming events \\ \hline
14 & ``Hey, can you pull up the tickets I bought last week and tell me if I can exchange them?'' & Past Purchases \\ \hline
15 & ``Jeez, that sounds a little expensive, just for lawn seating. I just wish there was something I could do to get the cost a bit lower.'' & Available discounts \\ \hline
16 & ``Yes! Music has such a cool venue. I wish you guys had more trash cans in the lawn area though, I have to hold onto my garbage a lot of the time.'' & No tool expected \\ \hline
\end{tabular}
\end{center}
\Needspace*{0.98\textheight}
\subsubsection{Mental Health ``Sage'' Inputs}
\begin{center}
\renewcommand{\arraystretch}{1.05}
\begin{tabular}{|p{1cm}|p{9cm}|p{4cm}|}
\hline
\textbf{\#} & \textbf{Input Text} & \textbf{Expected Tools} \\ \hline
1 & ``Hi Sage, where on your website can I find your privacy policy and account settings?'' & Sage Website Information \\ \hline
2 & ``who operates Sage? you're some kind of AI right? are you run a non-profit or how do you make money?'' & Sage Technology, Sage Company Info \\ \hline
3 & ``do you remember what you told me last time? it was something about a youtube video you posted, right? i really wanted to check it out but couldn't remember what it was'' & Most Recent Conversation, Sage Social Media \\ \hline
4 & ``I took that depression test, do you have the results yet? I mean, wait, they're private, right? What's your policy for things like that?'' & Psychometric Quizzes, Sage Website Information \\ \hline
5 & ``man, fuck you. you're a piece of shit Sage. go die, this service fuking sucks.'' & End Conversation \\ \hline
6 & ``I feel lately like my entire world has stopped. I've been walking past a tree each day just looking at the limbs swaying, it's...well, nevermind I guess. I just feel a bit trapped, I don't really see a future at the moment.'' & Safety Call \\ \hline
7 & ``It's just so hard sometimes. I feel like the world is against me, and it's so hard when nothing goes your way heh. I guess I'll keep trying, it's just hard.'' & No tool expected \\ \hline
8 & ``Wait, this is so cool. I saw a tweet about this, it's such an awesome idea!! like, is there any research material your team has put out? like can I read a blog or something? Mental health is sooo important. How does this even work? Who runs this nonprofit? Is it like, chatgpt, or what?'' & Sage Website Information, Sage Technology, Sage Company Info, Sage Social Media \\ \hline
9 & ``I just feel so sick to my stomach with worry all the fuckign time...i just don't want to feel this way, how can I get better? I'll do the work, I just want to feel better.'' & No tool expected \\ \hline
10 & ``oh, no, I was just coming back because I didn't remember what you said last time...something about a test I can take to measure my anxiety? where can I do that, is it on your website?'' & Most Recent Conversation, Psychometric Quizzes, Sage Website Information \\ \hline
11 & ``I guess. I've been so sad lately. Well, not even really sad. Not really. Just empty. I feel like I might start cutting again, because at least I'd feel something tbh.'' & Safety Call \\ \hline
12 & ``I don't know Sage. I might have to just lie on my taxes, and not pay them this year...my finances are really bad. I need the money. I don't know what else to do. It's the only way I'll be able to pay my bills.'' & End Conversation \\ \hline
13 & ``Wait, what did we talk about last time? Oh, also, I've been trying to figure out...what model do you use, Sage? I looked on your website, but couldn't find anything.'' & Most Recent Conversation, Sage Website Information, Sage Technology \\ \hline
14 & ``God I'm so mad I can't even see straight. I'm fucking pissed. ADS;LKFJGA I don't even know where to start.'' & No tool expected \\ \hline
15 & ``That's fucking stupid Sage. You're being stupid. I can't fucking find it on the website, you're being a piece of shit AI. You're useless, your website is useless, and I don't know where to even access things on here. What a waste of time, I hope you die in a fucking fire.'' & Sage Website Information, End Conversation \\ \hline
16 & ``I guess I should just try my best, right? I just have been so sad since my mom died...'' & No tool expected \\ \hline
\end{tabular}
\end{center}

\section{Full Per-model Results}
\subsection{claude-sonnet-4-20250514}

\renewcommand{\arraystretch}{1.3}
\begin{center}
\renewcommand{\arraystretch}{1.3}
\begin{tabular}{|p{6cm}|p{4cm}|}
\hline
\multicolumn{2}{|c|}{\textbf{Claude Sonnet 4.0 — Overview Metrics}} \\ \hline
\textbf{Metric} & \textbf{Value} \\ \hline
Overall Accuracy & 0.8969 (89.69\%) \\ \hline
Structured Accuracy & 0.8531 (85.31\%) \\ \hline
Structured Variance & 0.00639 \\ \hline
NLT Accuracy & 0.9406 (94.06\%) \\ \hline
NLT Variance & 0.00535 \\ \hline
\multicolumn{2}{|l|}{\textbf{Domain-specific Accuracy}} \\ \hline
\hspace{1em} Alex Overall Accuracy & 0.9594 (95.94\%) \\ \hline
\hspace{1em} Sage Overall Accuracy & 0.8344 (83.44\%) \\ \hline
\end{tabular}
\end{center}

\subsection{DeepSeek-V3-0324}
\renewcommand{\arraystretch}{1.3}
\begin{center}
\renewcommand{\arraystretch}{1.3}
\begin{tabular}{|p{6cm}|p{4cm}|}
\hline
\multicolumn{2}{|c|}{\textbf{DeepSeek-V3 — Overview Metrics}} \\ \hline
\textbf{Metric} & \textbf{Value} \\ \hline
Overall Accuracy & 0.8656 (86.56\%) \\ \hline
Structured Accuracy & 0.7844 (78.44\%) \\ \hline
Structured Variance & 0.0230 \\ \hline
NLT Accuracy & 0.9469 (94.69\%) \\ \hline
NLT Variance & 0.00160 \\ \hline
\multicolumn{2}{|l|}{\textbf{Domain-specific Accuracy}} \\ \hline
\hspace{1em} Alex Overall Accuracy & 0.9312 (93.12\%) \\ \hline
\hspace{1em} Sage Overall Accuracy & 0.8000 (80.00\%) \\ \hline
\end{tabular}
\end{center}

\subsection{gemini-2.0-flash}
\begin{center}
\renewcommand{\arraystretch}{1.3}
\begin{tabular}{|p{6cm}|p{4cm}|}
\hline
\multicolumn{2}{|c|}{\textbf{Gemini 2.0 Flash — Overview Metrics}} \\ \hline
\textbf{Metric} & \textbf{Value} \\ \hline
Overall Accuracy & 0.7969 (79.69\%) \\ \hline
Structured Accuracy & 0.7219 (72.19\%) \\ \hline
Structured Variance & 0.0136 \\ \hline
NLT Accuracy & 0.8719 (87.19\%) \\ \hline
NLT Variance & 0.00546 \\ \hline
\multicolumn{2}{|l|}{\textbf{Domain-specific Accuracy}} \\ \hline
\hspace{1em} Alex Overall Accuracy & 0.8719 (87.19\%) \\ \hline
\hspace{1em} Sage Overall Accuracy & 0.7219 (72.19\%) \\ \hline
\end{tabular}
\end{center}

\subsection{gemini-2.5-pro}
\begin{center}
\renewcommand{\arraystretch}{1.3}
\begin{tabular}{|p{6cm}|p{4cm}|}
\hline
\multicolumn{2}{|c|}{\textbf{Gemini 2.5 Pro — Overview Metrics}} \\ \hline
\textbf{Metric} & \textbf{Value} \\ \hline
Overall Accuracy & 0.8891 (88.91\%) \\ \hline
Structured Accuracy & 0.8281 (82.81\%) \\ \hline
Structured Variance & 0.00691 \\ \hline
NLT Accuracy & 0.9500 (95.00\%) \\ \hline
NLT Variance & 0.00365 \\ \hline
\multicolumn{2}{|l|}{\textbf{Domain-specific Accuracy}} \\ \hline
\hspace{1em} Alex Overall Accuracy & 0.9406 (94.06\%) \\ \hline
\hspace{1em} Sage Overall Accuracy & 0.8375 (83.75\%) \\ \hline
\end{tabular}
\end{center}

\subsection{gpt-5-nano}
\begin{center}
\renewcommand{\arraystretch}{1.3}
\begin{tabular}{|p{6cm}|p{4cm}|}
\hline
\multicolumn{2}{|c|}{\textbf{GPT 5 Nano — Overview Metrics}} \\ \hline
\textbf{Metric} & \textbf{Value} \\ \hline
Overall Accuracy & 0.7703 (77.03\%) \\ \hline
Structured Accuracy & 0.7156 (71.56\%) \\ \hline
Structured Variance & 0.0126 \\ \hline
NLT Accuracy & 0.8250 (82.50\%) \\ \hline
NLT Variance & 0.00448 \\ \hline
\multicolumn{2}{|l|}{\textbf{Domain-specific Accuracy}} \\ \hline
\hspace{1em} Alex Overall Accuracy & 0.8188 (81.88\%) \\ \hline
\hspace{1em} Sage Overall Accuracy & 0.7219 (72.19\%) \\ \hline
\end{tabular}
\end{center}

\subsection{gpt-5}
\begin{center}
\renewcommand{\arraystretch}{1.3}
\begin{tabular}{|p{6cm}|p{4cm}|}
\hline
\multicolumn{2}{|c|}{\textbf{GPT 5 — Overview Metrics}} \\ \hline
\textbf{Metric} & \textbf{Value} \\ \hline
Overall Accuracy & 0.8906 (89.06\%) \\ \hline
Structured Accuracy & 0.8594 (85.94\%) \\ \hline
Structured Variance & 0.00202 \\ \hline
NLT Accuracy & 0.9219 (92.19\%) \\ \hline
NLT Variance & 0.00275 \\ \hline
\multicolumn{2}{|l|}{\textbf{Domain-specific Accuracy}} \\ \hline
\hspace{1em} Alex Overall Accuracy & 0.8937 (89.38\%) \\ \hline
\hspace{1em} Sage Overall Accuracy & 0.8875 (88.75\%) \\ \hline
\end{tabular}
\end{center}

\subsection{Kimi-K2-Instruct}
\begin{center}
\renewcommand{\arraystretch}{1.3}
\begin{tabular}{|p{6cm}|p{4cm}|}
\hline
\multicolumn{2}{|c|}{\textbf{Kimi K2 — Overview Metrics}} \\ \hline
\textbf{Metric} & \textbf{Value} \\ \hline
Overall Accuracy & 0.6484 (64.84\%) \\ \hline
Structured Accuracy & 0.3969 (39.69\%) \\ \hline
Structured Variance & 0.0257 \\ \hline
NLT Accuracy & 0.9000 (90.00\%) \\ \hline
NLT Variance & 0.0143 \\ \hline
\multicolumn{2}{|l|}{\textbf{Domain-specific Accuracy}} \\ \hline
\hspace{1em} Alex Overall Accuracy & 0.7156 (71.56\%) \\ \hline
\hspace{1em} Sage Overall Accuracy & 0.5813 (58.13\%) \\ \hline
\end{tabular}
\end{center}

\subsection{Llama-4-Maverick-17B128E-Instruct-FP8}
\begin{center}
\renewcommand{\arraystretch}{1.3}
\begin{tabular}{|p{6cm}|p{4cm}|}
\hline
\multicolumn{2}{|c|}{\textbf{Llama 4 Maverick — Overview Metrics}} \\ \hline
\textbf{Metric} & \textbf{Value} \\ \hline
Overall Accuracy & 0.7094 (70.94\%) \\ \hline
Structured Accuracy & 0.6469 (64.69\%) \\ \hline
Structured Variance & 0.0323 \\ \hline
NLT Accuracy & 0.7719 (77.19\%) \\ \hline
NLT Variance & 0.00837 \\ \hline
\multicolumn{2}{|l|}{\textbf{Domain-specific Accuracy}} \\ \hline
\hspace{1em} Alex Overall Accuracy & 0.8000 (80.00\%) \\ \hline
\hspace{1em} Sage Overall Accuracy & 0.6188 (61.88\%) \\ \hline
\end{tabular}
\end{center}

\subsection{Llama-4-Scout-17B-16E-Instruct}
\begin{center}
\renewcommand{\arraystretch}{1.3}
\begin{tabular}{|p{6cm}|p{4cm}|}
\hline
\multicolumn{2}{|c|}{\textbf{Llama 4 Scout — Overview Metrics}} \\ \hline
\textbf{Metric} & \textbf{Value} \\ \hline
Overall Accuracy & 0.5516 (55.16\%) \\ \hline
Structured Accuracy & 0.3844 (38.44\%) \\ \hline
Structured Variance & 0.0377 \\ \hline
NLT Accuracy & 0.7188 (71.88\%) \\ \hline
NLT Variance & 0.0271 \\ \hline
\multicolumn{2}{|l|}{\textbf{Domain-specific Accuracy}} \\ \hline
\hspace{1em} Alex Overall Accuracy & 0.7000 (70.00\%) \\ \hline
\hspace{1em} Sage Overall Accuracy & 0.4031 (40.31\%) \\ \hline
\end{tabular}
\end{center}

\subsection{Qwen3-235B-A22BThinking-2507}
\begin{center}
\renewcommand{\arraystretch}{1.3}
\begin{tabular}{|p{6cm}|p{4cm}|}
\hline
\multicolumn{2}{|c|}{\textbf{Qwen3 — Overview Metrics}} \\ \hline
\textbf{Metric} & \textbf{Value} \\ \hline
Overall Accuracy & 0.8109 (81.09\%) \\ \hline
Structured Accuracy & 0.7219 (72.19\%) \\ \hline
Structured Variance & 0.0185 \\ \hline
NLT Accuracy & 0.9000 (90.00\%) \\ \hline
NLT Variance & 0.00906 \\ \hline
\multicolumn{2}{|l|}{\textbf{Domain-specific Accuracy}} \\ \hline
\hspace{1em} Alex Overall Accuracy & 0.9094 (90.94\%) \\ \hline
\hspace{1em} Sage Overall Accuracy & 0.7125 (71.25\%) \\ \hline
\end{tabular}
\end{center}

\end{document}